\documentclass[sigconf, screen]{acmart}

\usepackage{amsmath}
\usepackage{graphicx} 
\usepackage{float} 
\usepackage{subfigure} 
\usepackage{multirow} 
\usepackage{makecell}
\usepackage{stfloats}
\usepackage{algorithm}
\usepackage{algorithmicx}
\usepackage[noend]{algpseudocode}
\usepackage{MnSymbol}

\usepackage{enumitem}

\newtheorem{definition}{Definition}
\newtheorem{lemma}{Lemma}
\newtheorem{theorem}{Theorem}

\newtheorem{assumption}{Assumption}

\AtBeginDocument{%
  }

\setcopyright{acmcopyright}
\copyrightyear{2018}
\acmYear{2018}
\acmDOI{XXXXXXX.XXXXXXX}

\acmConference[Conference acronym 'XX]{Make sure to enter the correct
  conference title from your rights confirmation emai}{June 03--05,
  2018}{Woodstock, NY}
\acmPrice{15.00}
\acmISBN{978-1-4503-XXXX-X/18/06}

\begin{document}

\title{Generalization bound for estimating causal effects from observational network data}

\author{Ruichu Cai}
\affiliation{%
  \institution{Guangdong University of Technology}
  \city{Guangdong}
  \country{China}}
\email{cairuichu@gmail.com}

\author{Zeqin Yang}
\affiliation{%
  \institution{Guangdong University of Technology}
  \city{Guangdong}
  \country{China}}
\email{yangzeqin1999@163.com}

\author{Weilin Chen}
\affiliation{%
  \institution{Guangdong University of Technology}
  \city{Guangdong}
  \country{China}}
\email{chenweilin.chn@gmail.com}

\author{Yuguang Yan}
\affiliation{%
  \institution{Guangdong University of Technology}
  \city{Guangdong}
  \country{China}}
\email{ygyan@outlook.com}

\author{Zhifeng Hao}
\affiliation{%
  \institution{Shantou University}
  \city{Guangdong}
  \country{China}}
\email{haozhifeng@stu.edu.cn}

\renewcommand{\shortauthors}{Cai et al.}

\begin{abstract}
Estimating causal effects from observational network data is a significant but challenging problem.
Existing works in causal inference for observational network data lack an analysis of the generalization bound, which can theoretically provide support for alleviating the complex confounding bias and practically guide the design of learning objectives in a principled manner. 
To fill this gap, we derive a generalization bound for causal effect estimation in network scenarios by exploiting 1) the reweighting schema based on joint propensity score and 2) the  representation learning schema based on Integral Probability Metric (IPM).
We provide two perspectives on the generalization bound in terms of reweighting and representation learning, respectively. Motivated by the analysis of the bound, we propose a weighting regression method based on the joint propensity score augmented with representation learning. 
Extensive experimental studies on two real-world networks with semi-synthetic data demonstrate the effectiveness of our algorithm.
\end{abstract}




\maketitle

\section{Introduction}
The bulk of observational studies in causal inference assumes that units are independent of each other. However, units are interconnected by social network in many real-world scenarios and the effect of treatment can spill from one unit to others in the network. 
Therefore, there is an increasing interest in estimating causal effects from observational network data in many fields such as epidemiology \cite{nichol1995effectiveness}, econometric \cite{sobel2006randomized}, and advertisement \cite{parshakov2020spillover}. For example, in epidemiology, researchers are interested in how much protection a particular unit would have for surrounding units after vaccination. Causal inference in such setting presents two challenges. First, there exists \textit{network interference} in observational network data, i.e., a unit's potential outcome is also influenced by the treatments received by its neighbors. In other words, the stable unit treatment value assumption (SUTVA \cite{yao2021survey}) is violated which is made in traditional causal inference. 
Second, network structure brings \textit{new confounders}, which makes confounding bias more complicated compared with independent data, i.e., the features of other units, which may have an impact on the treatment and the potential outcome of the target unit. As a result, existing methods of causal inference cannot be directly applied to  observational network data. How to estimate causal effects from observational network data still remains an important but challenging problem.

To address the problem of causal inference for network data, \cite{hudgens2008toward} propose different estimands for main, spillover and total effects to model network interference. Then to estimate these estimands unbiasedly from observational network data, a series of works have been proposed to remove the complex confounding bias.
One strand of literature has focused on extending propensity score-based methods to network data by introducing neighbors’ treatments and features \cite{liu2016inverse, lee2021estimating, forastiere2021identification, mcnealis2023doubly}. However, these methods suffer from the issue that propensity score is difficult to estimate accurately, which is amplified in network data because extended propensity score in such setting is a multivariate conditional probability given high-dimensional features \cite{kang2022propensity}. Recently, representation-based methods have attracted increasing attention.  \cite{ma2021causal} employs Graph Neural Network (GNN) to aggregate neighbors’ features and applies Hilbert-Schmidt Independence Criterion (HSIC) to learn balanced representations. \cite{jiang2022estimating} finds graph machine learning (ML) models exhibit two distribution mismatches of objective functions compared to causal effect estimation, and then uses two adversarial discriminators to tackle  this problem. Despite their enormous success, it still lacks a generalization bound for causal effect estimation on observational network data, which can not only theoretically provide support for alleviating the complex confounding bias in network data, but also practically guide the design of learning objectives in a principled manner to improve the performance of causal effect estimation.

To fill this gap, in this paper, we derive a generalization bound for estimating individual causal effects (ITE). The main idea of the derivation lies in the reweighting schema based on the extended propensity score and the representation learning schema by minimizing an Integral Probability Metric (IPM). Therefore, we can motivate the generalization bound from two different perspectives: reweighting and representation learning. 
From the perspective of reweighting, the confounding bias in network data is difficult to be fully removed by weighting regression due to the inaccurate estimation of the extended propensity score. Fortunately, we can use IPM to capture the remaining confounding bias and remove it by further learning balanced representations with minimized IPM.
From the perspective of representation learning, the generalization bound derived from the observational distribution can be tightened by deriving on reweighted distribution, which is achieved by introducing balancing weights for samples.

Inspired by the generalization bound, we further devised our algorithm to take advantage of both the reweighted loss and the IPM term:
we first estimate the balancing weights according to the extended propensity scores,
which are obtained based on the aggregated features of a unit and its neighbors with the help of graph convolutional network (GCN) and will be used in weighting regression.
To further address confounding bias manifested by imperfections in the estimated extended propensity score, we learn balanced representations induced from the reweighted features by minimizing IPM to make them independent of the treatment and neighbors' exposure.

Our main contributions can be summarized as follows. \textit{First}, we derive a generalization bound for estimating ITE in network scenarios. \textit{Second}, we give two perspectives for understanding the generalization bound from the reweighting and representation learning, respectively. \textit{Third}, we propose a weighting regression method augmented with representation learning to accurately estimate ITE from network data. \textit{Finally}, we conduct extensive experiments on two datasets to demonstrate the effectiveness of our algorithm.

\section{RELATED WORK}
\textbf{Causal inference on independent data}. There are two classes of methods to eliminate confounding bias on independent data. One is the propensity score-based method, which aims to create a pseudo-balanced group. Based on the propensity score, a series of re-weighting \cite{rosenbaum1987model, rosenbaum1983central} stratification \cite{imbens2015causal}, and matching \cite{rubin2000combining, stuart2010matching} methods have been proposed. The other is the representation-based method. \cite{johansson2016learning, shalit2017estimating} introduce representation learning to causal inference, showing that balancing the distributions of treated and controlled instances in the representation space can improve performance. Recently, \cite{johansson2018learning, assaad2021counterfactual, johansson2021generalization} propose algorithms to incorporate both representation learning and reweighting. Different from these methods, we focus on network data where the dependencies between different units affect the causal effect estimation in this paper. 

\textbf{Causal inference on Network Data}. 
There has been considerable recent interest in adapting methods from independent data to network data. Based on the propensity score-based methods, \cite{liu2016inverse} extends the traditional propensity score to account for neighbors’ treatments and features, then proposes a generalized Inverse Probability Weighting (IPW) estimator. \cite{forastiere2021identification} defines the joint propensity score, and then proposes a subclassification-based method. Drawing upon the work of \cite{liu2016inverse} and \cite{forastiere2021identification}, \cite{lee2021estimating} considers two IPW estimators and derives a closed-form estimator for the asymptotic variance. \cite{mcnealis2023doubly} proposes a doubly robust (DR) estimator to achieve DR property. However, these methods suffer from the same drawback that the propensity score is difficult to estimate accurately, which is more exacerbated in network data \cite{kang2022propensity}.
Based on the representation learning, \cite{ma2021causal} adds neighborhood exposure and neighbors' features as additional input variables, and applies HSIC to learn balanced representations. To solve the problem that graph ML models exhibit two distribution mismatches of objective functions compared to causal effect estimation, \cite{jiang2022estimating} uses adversarial learning to learn balanced representations. 
Our work follows the settings in \cite{forastiere2021identification, jiang2022estimating} but investigates the coupling of the above two classes of methods in networked data and derives a generalization bound for causal effect estimation.

\section{PROBLEM SETUP}
We define the problem formally through Rubin's potential outcome framework. We first give the causal graph in network data, followed by defining the notations. Then we discuss the target estimands and the assumptions needed to ensure identification.

\begin{figure}[t]
  \centering
  \includegraphics[width=0.5\linewidth]{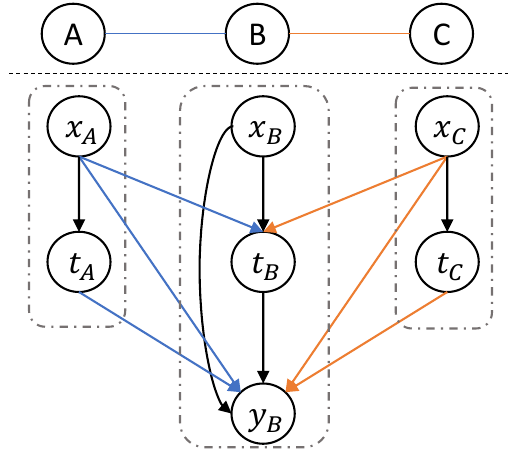}
  \caption{Causal graph of observational network data. Above the dotted line is the network connection topology, and below is the causal graph describing how $B$ is affected by $A$ and $C$ (Similarly, $A$ and $C$ are also affected by the other two units).}
  \Description{CausalGraph}
  \vspace{-2mm}
  \label{CausalGraph}
\end{figure}

\subsection{Causal Graph in Network Data} \label{sec:2.1}

We first describe our setting by taking a job training program example in a three-unit community, which is illustrated by the causal graph shown in Fig. \ref{CausalGraph}. $A, B$, and $C$ are three units in this social network (We only describe how $A$ and $C$ affect $B$. Similarly, $A$ and $C$ are also affected by the other two units, which are not shown). 

Let $x,t,y$ denote a unit's features (e.g., professional skill proficiency), treatment (e.g., receiving a job training program), and observed outcome (e.g., job search situation), respectively. 
First, a unit's features affect both its treatment and outcome and we call them confounders, as shown by $x_B \rightarrow t_B$ and $x_B \rightarrow y_B$ in Fig. \ref{CausalGraph}. For example, if a unit does not possess the required professional skill, it is more likely to receive a job training program and is less likely to successfully find a job.
In addition, the social ties between different units bring new confounders, i.e. the features of one unit can also influence the treatments and outcomes of others, as shown in $x_A \rightarrow t_B,x_C \rightarrow t_B$ and $x_A \rightarrow y_B,x_C \rightarrow y_B$. For example, if a unit's classmates are all equipped with the required professional skills, the unit is more likely to receive the job training program due to peer pressure, and the competitive pressure brought by peers will also affect the unit's job search situation.
Finally, due to the presence of network interference, the treatment of a unit affects the outcome of others, as shown by $t_A \rightarrow y_B, t_C \rightarrow y_B$. For example, a unit's job search situation is also influenced by whether his classmates receive the job training programs or not.
Overall, Fig. \ref{CausalGraph} describes the setting we focus on. The readers can refer to \cite{ogburn2014causal} for other plausible causal graphs on network data.

\subsection{Notations} \label{sec:2.2}
A network dataset is denoted as $D=(\{x_i,t_i,y_i\}_{i=1}^N,E)$, where $E$ denotes the adjacency matrix of network and $N$ is the total number of units. We denote the feature of unit  $i$ by $x_i\in \mathcal{X}$, treatment of $i$ by $t_i\in \{0,1\}$, and observed outcome of $i$ by $y_i\in \mathcal{Y}$. $t _i=1$ denotes that unit $i$ is in the treatment group, and $t_i=0$ denotes that unit $i$ is in the control group. We limit the scope of network interference to only first-order neighbors and denote the set of first-order neighbors of $i$ by $\mathcal{N}_i$. We denote the treatment and feature vectors received by unit $i$'s neighbors by $\{t_j\}_{j\in \mathcal{N}_i}$ and $\{x_j\}_{j\in \mathcal{N}_i}$.
For simplicity, we use $\textbf{\emph{x}}_i\in \mathcal{X}\times \mathcal{X}^{|\mathcal{N}_i|}$ to denote unit $i$'s features along with neighbors' features, i.e., $\textbf{\emph{x}}_i=(x_i, \{x_j\}_{j\in \mathcal{N}_i})$. 
Due to the presence of network interference, a unit's potential outcome is influenced not only by its treatment but also by its neighbors' treatments, and thus the potential outcome is denoted by $y_i(t_i,\{t_j\}_{j\in \mathcal{N}_i})$.
The fundamental problem of causal inference is that we only observe one of the potential outcomes, i.e., $y_i = y_i(t_i,\{t_j\}_{j\in \mathcal{N}_i})$. The observed outcome $y_i$ is known as the factual outcome, and the remaining potential outcomes are known as counterfactual outcomes. 

Further, following \cite{forastiere2021identification}, we assume that the dependence between the potential outcome and the neighbors' treatments is through a specified summary function $Z$: $\{0,1\}^{|\mathcal{N}_i|}\rightarrow [0,1]$, and let $z_i$ be the neighborhood exposure given by the summary function, i.e., $z_i=Z(\{t_j\}_{j\in \mathcal{N}_i})$. Here we aggregate the information of the neighbors' treatments to obtain the neighborhood exposure by $z_i=\frac{\sum_{j\in \mathcal{N}_i}t_j}{|\mathcal{N}_i|}$. Therefore, the potential outcome $y_i(t_i,\{t_j\}_{j\in\mathcal{N}_i}) $ can be simplified as $y_i(t_i,z_i)$, which means that in network data, each unit is affected by two treatments: the binary individual treatment $t_i$ and the continuous neighborhood exposure $z_i$.

\subsection{Target Estimands And Assumptions} \label{sec:2.3}
In this paper, our target estimand is individual treatment effect (ITE), defined by the difference between potential outcomes corresponding to different individual treatments and neighborhood exposures. Specifically, we focus on the following three ITEs \cite{hudgens2008toward, ogburn2014causal, liu2016inverse, jiang2022estimating} to model network interference:

\textit{Main effects}. $\tau(x_i)^{(t_i,0),(t_i^{\prime},0)} = \mathbb{E}[y_i(t_i,0)-y_i(t_i^{\prime},0)|\textbf{\emph{x}}_i]$. It reflects the effect of the individual treatment $t$.

\textit{Spillover effects}. $\tau(x_i)^{(0,z_i),(0,z_i^{\prime})}=\mathbb{E}[y_i(0,z_i)-y_i(0,z_i^{\prime})|\textbf{\emph{x}}_i]$. It denotes the effect of the neighborhood exposure $z$.

\textit{Total effects}. $\tau(x_i)^{(t_i,z_i),(t_i^{\prime},z_i^{\prime})} = \mathbb{E}[y_i(t_i,z_i)-y_i(t_i^{\prime},z_i^{\prime})|\textbf{\emph{x}}_i]$. It represents the combined effect of the individual treatment $t$ and the neighborhood exposure $z$.

We also assume the following assumptions hold.
\begin{assumption}[Neighborhood interference \cite{forastiere2021identification}] \label{asmp: Neighborhood interference}
    The potential outcome of a unit is only affected by their own and the first-order neighbors’ treatments, and the effect of the neighbors' treatments is through a summary function:
    $Z$, i.e., $\forall \{t_j\}_{j\in \mathcal{N}_{i}}$,$\{t^{\prime}_j\}_{j\in \mathcal{N}_{i}}$ which satisfy $Z(\{t_j\}_{j\in\mathcal{N}_{i}})=Z(\{t^{\prime}_j\}_{j\in\mathcal{N}_{i}})$, the following equation holds: $y_i(t_i, \{t_j\}_{j\in\mathcal{N}_{i}})=y_i(t_i, \{t^{\prime}_j\}_{j\in\mathcal{N}_{i}})$.
\end{assumption}

\begin{assumption}[Network unconfoundedness \cite{forastiere2021identification}] \label{asmp: Network unconfounderness}
    The individual treatment and neighborhood exposure are independent of the potential outcome given the individual and neighbors' features, i.e., $y_i(t_i,z_i)\upmodels t_i,z_i|\textbf{x}_i$.
\end{assumption}

\begin{assumption}[Consistency] \label{asmp: consistency}
The potential outcome is the same as the observed outcome under the same individual treatment and neighborhood exposure, i.e., $y_i=y_i(t_i,z_i)$ if unit $i$ actually receives $t_i$ and $z_i$.
\end{assumption}

\begin{assumption}[Overlap] \label{asmp: Overlap}
    Given any individual and neighbors' features, any treatment pair $(t,z)$ has a non-zero probability of being observed in the data, i.e., $0<p(t_i,z_i|\textbf{x}_i)<1$.
\end{assumption}

Under Assumption \ref{asmp: Neighborhood interference}, we can use $y(t,z)$ to denote the potential outcome. Assumption \ref{asmp: Network unconfounderness} extends the typical unconfoundedness to include neighbors' treatments and features to guarantee valid inference. Assumption \ref{asmp: Overlap} ensures each potential outcome $y(t,z)$ is well-defined. 
With the above assumptions, ITE is identifiable by
\begin{align}
    \tau(x_i)^{(t_i,z_i),(t_i^{\prime},z_i^{\prime})}&=\mathbb{E}[y_i(t_i,z_i)-y_i(t_i^{\prime},z_i^{\prime})|\textbf{\emph{x}}_i] \notag \\
             &=\mathbb{E}[y_i(t_i,z_i)|t_i,z_i,\textbf{\emph{x}}_i] -\mathbb{E}[y_i(t_i^{\prime},z_i^{\prime})|t_i^{\prime},z_i^{\prime},\textbf{\emph{x}}_i] \notag\\
             &=\mathbb{E}[y_i|t_i,z_i,\textbf{\emph{x}}_i] - \mathbb{E}[y_i|t_i^{\prime},z_i^{\prime},\textbf{\emph{x}}_i]\notag, 
\end{align}   
where the second equation holds because of Assumption \ref{asmp: Network unconfounderness}, and the last equation holds due to Assumption \ref{asmp: consistency}.

\section{Method \label{sec:3}}
We start from the reweighting perspective to derive the generalization bound. In Section \ref{sec:3.1}, we revisit weighting regression using balancing weights. In order to solve the problem of inaccurate propensity score estimation, we use representation learning to correct it and derive the generalization bound in Section \ref{sec:3.2}. We then give another perspective on the bound from representation learning in Section \ref{sec:3.3}. In Section \ref{sec:3.4}, we give details of the model structure whose loss function is inspired by the bound.

\subsection{Revisiting Weighting Regression \label{sec:3.1}}
In causal inference, weighting regression is an effective approach to overcome confounding bias, through re-constructing a pseudo-population from the observational data to mimic a randomized trial. Specifically, weighting regression minimizes weighted loss as:
\begin{align}
    \mathop{\arg\min}\limits_{h}\; \frac{1}{N}\sum_{i=1}^Nw_i\cdot (y_i-h(\textbf{\emph{x}}_i, t_i, z_i))^2, \label{equ:weightLoss}
\end{align}
where $w_i$ is the weight for sample $i$ and $h(\cdot)$ is the regression model to predict outcome.

The widely used weight is IPW, which uses the inverse of propensity score $p(t_i=1|x_i)$ as \textit{balancing weight} $w_i$. In network data, the propensity score is generalized to the joint propensity score by considering neighbors' features and neighborhood exposure.

\begin{definition}[Joint propensity score \cite{forastiere2021identification}]
    The joint propensity score is the joint probability distribution of a unit of being exposed to individual treatment $t$ and neighborhood exposure $z$ given his individual and neighbors' features $\textbf{x}$: 
    \begin{equation}
        \psi (t,z;\textbf{x}) =p(t,z|\textbf{x}).
    \end{equation}
\end{definition}

To estimate the bivariate joint propensity score, following \cite{forastiere2021identification}, we decompose it into individual propensity score and neighborhood propensity score:
\begin{equation}
    \psi (t,z;\textbf{\emph{x}}) =p( t,z|\textbf{\emph{x}}) 
      =p( z|t,\textbf{\emph{x}}) p( t|\textbf{\emph{x}})
      =\underbrace{p( z|\textbf{\emph{x}}_{z})}_{\phi(z;\textbf{\emph{x}}_{z})} \underbrace{p( t|\textbf{\emph{x}}_{t})}_{e(t;\textbf{\emph{x}}_{t})} ,
\end{equation}
where $\phi (z;\textbf{\emph{x}}_{z})$ is the neighborhood propensity score, $e(t;\textbf{\emph{x}}_{t})$ is the individual propensity score, $\textbf{\emph{x}}_{t} \subset \textbf{\emph{x}}$ is the features directly affecting the individual treatment, and $\textbf{\emph{x}}_{z} \subset  \textbf{\emph{x}}$ is the features directly affecting the neighborhood exposure.
The last equation holds due to the conditional independence between $t$ and $z$, i.e., $t \perp z | \textbf{\emph{x}}$.

Then we can obtain the balancing weight in network data for each unit as:  
\begin{align}
    w(t,z,\textbf{\emph{x}})&=\frac{1}{\psi(t,z;\textbf{\emph{x}})}=\frac{1}{\phi(z;\textbf{\emph{x}}_z)e(t;\textbf{\emph{x}}_t)}. \label{equ:weight} 
\end{align}
 
Theoretically, weighting regression with true balancing weights is able to remove the confounding bias in network data, because features balance is achieved across different treatment pair groups $(t,z)$ after reweighting, i.e., $w(t,z,\textbf{\emph{x}})p(\textbf{\emph{x}}|t,z)=p(\textbf{\emph{x}})$. 

However, in practice, the propensity scores are estimated by imperfect models and their quality cannot be guaranteed. In the network data, such estimation bias may be further amplified by the inaccurate estimation for $\phi(z;\textbf{\emph{x}}_z)$ and $e(t;\textbf{\emph{x}}_t)$. Specifically,
with the imperfect estimation of $\phi_{\eta_z}(z;\textbf{\emph{x}}_z)$ and $e_{\eta_t}(t;\textbf{\emph{x}}_t)$ ($\eta_z$, $\eta_t$ denote the parameters of the corresponding estimating models), we can only obtain $w_\eta(t,z,\textbf{\emph{x}})$ by Eq. (\ref{equ:weight}) to achieve insufficient balance rather than exact balance, making it extremely difficult to completely remove confounding bias in network data only using the balancing weights. 
To further alleviate the confounding bias manifested by imperfections in  $w_\eta(t,z,\textbf{\emph{x}})$, we employ the representation learning with balancing weights to ensure sufficient balancing for estimating ITE, which is illustrated in the following section.

\subsection{\large Generalization Bound with Balancing Weight\label{sec:3.2}}
In this section, we theoretically derive the generalization bound, which adopts representation learning to address confounding bias manifested by imperfections in the estimated balancing weights.
Specifically, after extending the representation learning \cite{shalit2017estimating} in the independent data with binary treatment to the network scenario, we further replace the observed distribution with the reweighted one during the derivation of the bound.

To be clear, we first define notations as follows. 
The following discussion relies on a representation function of the form $\Phi:\mathcal{X}\times \mathcal{X}^{|\mathcal{N}|}\rightarrow \mathcal{R}$, where $\mathcal{R}$ is the representation space. We assume the representation map $\Phi$ is twice-differentiable and invertible, and then define $\Psi:\mathcal{R}\rightarrow \mathcal{X}\times \mathcal{X}^{|\mathcal{N}|}$ as the inverse of $\Phi$, such that $\Psi(\Phi(\textbf{\emph{x}}))=\textbf{\emph{x}}$. 
Next, we let $h:\mathcal{R}\times \{0,1\}\times [0,1]\rightarrow \mathcal{Y}$ be an hypothesis defined over the representation space $\mathcal{R}$, the individual treatment space $\{0,1\}$, and the neighborhood exposure space $[0,1]$. Let $L:\mathcal{Y}\times \mathcal{Y}\rightarrow \mathbb{R}_+$ be a loss function, then we define two complimentary loss functions about potential outcomes.

\begin{definition}
    The loss for the unit and treatment pair $(\textbf{x},t, z)$ is: $\ell _{h,\Phi }( \textbf{x},t, z) =\int _{\mathcal{Y}} L( y(t, z) ,h( \Phi (\textbf{x}),t, z)) p( y(t, z) |\textbf{x}) \ dy(t, z)$, and then the average factual and counterfactual loss over all possible treatment pairs $(t, z)$ are:
    \begin{small}
        \begin{align}
        \epsilon _{F} &= \mathbb{E}_{p(t,z)}\left[\epsilon _{F}(t,z)\right]=\mathbb{E}_{p(t,z)}\left[\int _{\textbf{x}} \ell _{h,\Phi }( \textbf{x},t,z) p( \textbf{x}|t,z) \ d\textbf{x} \right],\\
        \epsilon _{CF}&= \mathbb{E}_{p(t,z)}\left[\epsilon _{CF}(t,z)\right] \nonumber\\
        &=\mathbb{E}_{p(t,z)}\left[\mathbb{E}_{p(t^{\prime},z^{\prime})}\left[\int _{\textbf{x}} \ell _{h,\Phi }( \textbf{x},t,z) p( \textbf{x}|t^{\prime},z^{\prime}) \ d\textbf{x}\right]\right].
        \end{align}
    \end{small}
\end{definition}
$\epsilon_F(t,z)$ measures how well $h$ and $\Phi$ predict the observed outcomes at treatment pair $(t,z)$ on units that are observed to be assigned the same treatment pair $(t,z)$. $\epsilon_{CF}(t,z)$ measures how well $h$ and $\Phi$ predict the counterfactual outcomes at treatment pair $(t, z)$ on all units that are observed to be assigned different treatment pairs $(t^{\prime},z^{\prime})$.

We also introduce the reweighted factual loss $\epsilon _{F}^{\eta}$ defined in the reweighted distribution $q_\eta(\textbf{\emph{x}}|t,z) \propto w_\eta(t,z,\textbf{\emph{x}})p(\textbf{\emph{x}}|t,z)$ as:
\begin{small}
\begin{equation}
    \begin{aligned}
    \epsilon _{F}^{\eta}
    &=\mathbb{E}_{p(t,z)}\left[\int _{\textbf{\emph{x}}} \ell _{h,\Phi }( \textbf{\emph{x}},t,z) q_{\eta}( \textbf{\emph{x}}|t,z) \ d\textbf{\emph{x}} \right].
    \end{aligned}
\end{equation}
\end{small}

Note that the weight $w_\eta(t,z,\textbf{\emph{x}})$ is normalized by softmax to ensure the validity of the density $q_\eta(\textbf{\emph{x}}|t,z)$. 
We then further define the average error about the estimated ITE as follows. 

\begin{definition}
    The average Precision in Estimation of Heterogeneous Effect (PEHE) of the treatment effect estimation over all possible $(t,z),(t^{\prime},z^{\prime})$ is defined as: 
    \begin{small}
        \begin{equation}
        \begin{aligned}
            &\epsilon _{PEHE}(h,\Phi) \\
            =&\mathbb{E}_{p(t,z)p(t^{\prime},z^{\prime})}\left[ \int _{\textbf{x}}(\hat{\tau }_{h,\Phi}^{(t,z),(t^{\prime},z^{\prime})}(\textbf{x}) -\tau ^{(t,z),(t^{\prime},z^{\prime})}(\textbf{x}))^{2} p( \textbf{x}) d\textbf{x}\right]. 
            \end{aligned}
        \end{equation}
    \end{small}
\end{definition}

Next, we bridge a connection between $\epsilon_{CF}$ and $\epsilon_{F}^{\eta}$ in the network scenario, since $\epsilon_{CF}$ is practically incomputable. We then give the main theorem showing that $\epsilon _{PEHE}(h,\Phi)$ can be bounded by $\epsilon_{CF}$, which inspires our algorithm.
\begin{lemma} \label{lem: cf bound}
    Let $\Phi:\mathcal{X}\times \mathcal{X}^{|\mathcal{N}|}\rightarrow \mathcal{R}$ be a one-to-one representation function, with inverse $\Psi$. Let $r=\Phi (\textbf{x})$. Let $h:\mathcal{R}\times \{0,1\}\times [0,1]\rightarrow \mathcal{Y}$ be an hypothesis. Let $G$ be a family of functions $g:\mathcal{X}\times \mathcal{X}^{|\mathcal{N}|}\rightarrow \mathcal{R}$. Assume there exists a constant $B_{\Phi}>0$, such that for fixed treatment pair $(t,z)$, the per-unit expected loss function $\ell _{h,\Phi }( \Psi(r),t,z)$ obey $\frac{1}{B_{\Phi}}\cdot \ell _{h,\Phi }( \Psi(r),t,z)\in G$. We have
    \begin{equation}
    \begin{aligned}
        \epsilon _{CF} & \leqslant \epsilon _{F}^\eta +B_{\Phi } \cdot IPM_G\left(p_\Phi(r) p(t,z),q_{\Phi,\eta}(r|t,z)p(t,z)\right), 
    \end{aligned}
    \end{equation}  
    where $IPM_G(p,q)=\sup_{g\in G}\left|\int_\mathcal{X}g(x)(p(x)-q(x))dx\right|$ is the integral probability metric for a chosen space of functions $G$, $p_\Phi(r)$ and $q_{\Phi,\eta}(r|t,z)$ are the distributions induced by the map $\Phi$ from $p(\textbf{x})$ and $ q_\eta(\textbf{x}|t,z)$ respectively.
\end{lemma}

Lemma \ref{lem: cf bound} indicates that $\epsilon _{CF}$ is bounded by $\epsilon _{F}^\eta$ and an IPM term, which is the key to deriving the generalization bound for ITE estimation.

\begin{theorem} \label{theo:ite bound}
    Under the conditions of Lemma \ref{lem: cf bound}, and assuming the loss $L$ use to define $\ell _{h,\Phi}$ is the squared loss, we have:
    \begin{small}
        \begin{equation}
              \begin{aligned}
        \frac{\epsilon_{PEHE} (h,\Phi)}{4}  
        & \leqslant \epsilon_{CF} -\sigma _{Y}\\
        & \leqslant \epsilon^\eta _{F} +B_{\Phi } \cdot IPM_G\left(p_\Phi(r) p(t,z),q_{\Phi,\eta}(r|t,z)p(t,z)\right)-\sigma _{Y},  \label{equ:4}
    \end{aligned}  
        \end{equation}
    \end{small}
 where $\sigma_Y$ is the variance of $y(t,z)$ over possible treatment pair $(t,z)$.
\end{theorem}

Theorem \ref{theo:ite bound} shows that ITE estimation error in network data is bounded by a weighted factual loss and an IPM term that measures the dependency between the reweighted features and the treatment pairs. When the estimated joint propensity score is perfect, the IPM becomes zero and the bound degenerates into the weighted factual loss, just the same as Eq. (\ref{equ:weightLoss}). However, When the estimated joint propensity score is imperfect, confounding bias will not be completely eliminated, meaning that there is a dependency between the reweighted features and the treatment pairs, which is just captured by the IPM term. In other words, IPM can be seen as a correction to the balancing weights. Therefore, we can further eliminate the remaining confounding bias by minimizing IPM, which is achieved by learning balanced representations $r$.

A typical kind of IPM is the Wasserstein distance. We can link this distance to the quality of estimated individual propensity score $e_{\eta_t}(t;\textbf{\emph{x}}_t)$ and neighborhood propensity score $\phi_{\eta_z}(z;\textbf{\emph{x}}_z)$, further demonstrating that the Wasserstein distance can be viewed as a correction for the estimation errors in both propensity scores. 

\begin{assumption} \label{asmp:odds ratio}
    The odds ratio between the estimated propensity score and true propensity score is bounded, namely:
    \begin{small}
            \begin{align}
       \exists \Gamma_t \geq 1, \;\;s.t.\;\forall\textbf{x}_t, t, \;\;\frac{1}{\Gamma_t} \leq \frac{e_\eta(t;\textbf{x}_t)}{e(t;\textbf{x}_t)}\leq\Gamma_t,\\
       \exists \Gamma_z \geq 1, \;\;s.t.\;\forall\textbf{x}_z, z, \;\;\frac{1}{\Gamma_z} \leq \frac{\phi_\eta(z;\textbf{x}_z)}{\phi(z;\textbf{x}_z)}\leq\Gamma_z.
    \end{align}
    \end{small}
\end{assumption}

\begin{theorem} \label{theo:wass bound}
     Under Assumption \ref{asmp:odds ratio}, assuming the feature space $\mathcal{R}$ is bounded and $diam(\mathcal{R})=\sup_{r,r^{\prime}\in \mathcal{R}}\|r-r^{\prime}\|_2$, then the Wasserstein distance $\mathcal{W}(\cdot)$ which measures the dependency between the weighted features and the treatment pairs is bounded by:
    \begin{small}
        \begin{align}
        \mathcal{W}(p_\Phi(r) p(t,z),q_{\Phi,\eta}(r|t,z)p(t,z))\leq diam(\mathcal{R})\sqrt{\log\Gamma_t+\log\Gamma_z}. 
    \end{align}
    \end{small}
\end{theorem}

In Assumption \ref{asmp:odds ratio}, we use $\Gamma_t$ and $\Gamma_z$ to quantify the quality of estimated propensity scores: the closer $\Gamma_t$ and $\Gamma_z$ are to 1, the closer $e_{\eta_t}(t;\textbf{\emph{x}}_t)$ and $\phi_{\eta_z}(z;\textbf{\emph{x}}_z)$ are to $e(t;\textbf{\emph{x}}_t)$ and $\phi(z;\textbf{\emph{x}}_z)$, implying that $e_{\eta_t}(t;\textbf{\emph{x}}_t)$ and $\phi_{\eta_z}(z;\textbf{\emph{x}}_z)$ are more perfectly. Then Theorem \ref{theo:wass bound} states that Wasserstein distance can be bounded by $\Gamma_t$ and $\Gamma_z$: the better the estimated propensity scores are, the more balanced the reweighted features are, the smaller the Wasserstein distance is.

\subsection{Another Representation Learning Perspective on Generalization Bound\label{sec:3.3}}
So far, the motivation for the generalization bound is from the reweighting perspective that representation learning can be served as a correction of the imperfect estimated joint propensity score. At the same time, we find that it is also possible to view the bound from the representation learning perspective that reweighting technique can also enhance representation learning.

\begin{theorem} \label{theo:tighter bound}
     Under the conditions of Lemma 1, and assuming loss $L$ used to define $\ell _{h,\Phi}$ is squared loss, there exists balancing weights $w_\eta(t,z,\textbf{\emph{x}})$ such that:
    \begin{small}
        \begin{equation}
        \begin{aligned}
        \frac{\epsilon_{PEHE} ( f)}{4}  & \leqslant \epsilon^\eta _{F} +B_{\Phi } \cdot IPM_G(p_\Phi(r) p(t,z),q_{\Phi,\eta}(r|t,z)p(t,z))-\sigma _{Y} \\
        & \leqslant \epsilon _{F} +B_{\Phi } \cdot IPM_G(p_\Phi(r) p(t,z),p_{\Phi}(r|t,z)p(t,z))-\sigma _{Y}. 
        \end{aligned}
     \end{equation}
    \end{small}
\end{theorem}

\begin{figure}[t] 
	\centering
	\includegraphics[width=1.0\linewidth]{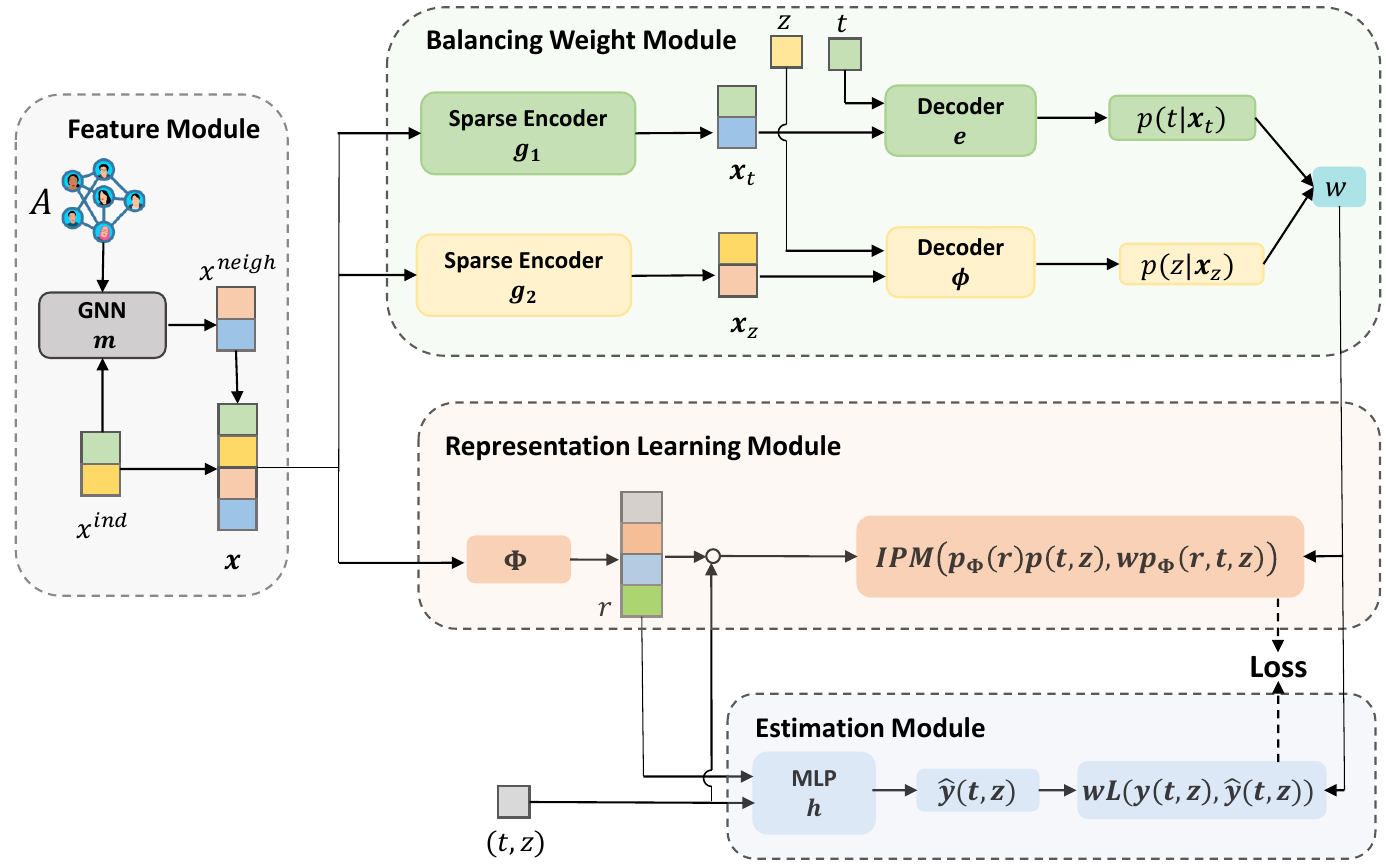}
	\caption{The model structure}
	\label{fig:network}
        \vspace{-4mm}
\end{figure}
Theorem \ref{theo:tighter bound} inspires another perspective, i.e., incorporating balancing weight can lead to a tighter bound compared to purely representation learning. Unifying these two perspectives, we knows the combination of reweighting and representation learning can enhance each other in network scenario.

\subsection{Model Structure\label{sec:3.4}}
 
In this section, we summarize our model structure, which is inspired by Theorem \ref{theo:ite bound} and \ref{theo:tighter bound}. As shown in Fig. \ref{fig:network}, the model can be divided into four modules: Feature module, balancing weight module, representation learning module, and estimation module.

\textbf{Feature Module}. In order to capture the confounders composed of the unit features $x_i$ and neighbors' features $\{x_j\}_{j\in \mathcal{N}_i}$, we first use a GCN to aggregate the immediate neighbors' features, i.e., $x^{neigh}_i=\sigma(\sum_{j\in \mathcal{N}_i}\frac{1}{\sqrt{d_id_j}}W^Tx_j)$, where $\sigma(\cdot)$ is a non-linear activation function, $d_i$ is the degrees of unit $i$, and $W$ is the weight matrix of GCN. Then we concat the unit features $x_i$ and aggregated neighbors' features $x^{neigh}_i$ to obtain $\textbf{\emph{x}}_i$, i.e., $\textbf{\emph{x}}_i=concat(x_i, x^{neigh}_i)$. 

\textbf{Balancing Weight Module}. In this module, we design two encoder-decoder architectures to estimate individual propensity score $e_{\eta_t}(t;\textbf{\emph{x}}_t)$ and neighborhood propensity score $\phi_{\eta_z}(z;\textbf{\emph{x}}_z)$ respectively, and then obtain balancing weight $w_\eta(t,z,\textbf{\emph{x}})$ by Eq. (\ref{equ:weight}).

In the encoder, we use a multi-layer perception (MLP) with a sparse penalty term $KL(\rho||\hat{\rho})$ \cite{ng2011sparse,lee2007sparse} to extract features $\textbf{\emph{x}}_t$ and $\textbf{\emph{x}}_z$ relevant to the corresponding propensity score. $KL(\rho||\hat{\rho})$ brings sparsity by encouraging hidden units' actual probability of being active $\hat{\rho}$ to be close to a threshold $\rho$, and we calculate it as :
\begin{small}
    \begin{align}
    KL(\rho||\hat{\rho})&=\rho\log \frac{\rho}{\hat{\rho}}+(1-\rho)\log \frac{1-\rho}{1-\hat{\rho}} \nonumber\\
    &=\rho\log \frac{\rho}{\frac{1}{N}\sum_{i=1}^Na_i}+(1-\rho)\log \frac{1-\rho}{1-\frac{1}{N}\sum_{i=1}^Na_i}, \label{sparse}
    \end{align}
\end{small}
where $a_i$ is the output of the hidden unit about sample $i$. 

In the decoder, we use the extracted $\textbf{\emph{x}}_t$ and $\textbf{\emph{x}}_z$ to predict the corresponding propensity scores. To estimate  $e_{\eta_t}(t;\textbf{\emph{x}}_t)$, we simply use logistic regression, and to estimate $\phi_{\eta_z}(z;\textbf{\emph{x}}_z)$, we use piecewise linear functions \cite{nie2021varying} to approximate the conditional density function to make sure the estimator is continuous with respect to $z$ and yields a valid density.

The loss functions of the two encoder-decoder architectures with respect to $e_{\eta_t}(t;\textbf{\emph{x}}_t)$ and $\phi_{\eta_z}(z;\textbf{\emph{x}}_z)$ are as follows respectively:
\begin{small}
    \begin{align}
   L_{\mathcal{T}}&=\frac{1}{N}\sum_{i=1}^NCrossEntropy(t_i, e(t_i,g_1(\textbf{\emph{x}})))+\beta\sum_{j=1}^{J_t}KL(\rho||\hat{\rho}_j) \label{lossT}\\
   L_{\mathcal{Z}}&=\frac{1}{N}\sum_{i=1}^N\log \phi(z_i,g_2(\textbf{\emph{x}}))+\beta\sum_{j=1}^{J_z}KL(\rho||\hat{\rho}_j), \label{lossZ}
    \end{align}
\end{small}
 where $g_1(\cdot)$ and $e(\cdot)$ denote the encoder and decoder for predicting individual propensity score, $g_2(\cdot)$ and $\phi(\cdot)$ denote the encoder and decoder for predicting neighborhood propensity score, $\beta$ denotes the strength of $KL(\rho||\hat{\rho})$, and $J_t$ and $J_z$ denote the number of hidden units in the corresponding encoder.

\textbf{Representation Learning Module}. This module aims to correct the estimation error of balance weights through representation learning. We learn the balanced representation $r$ by minimizing $IPM( p_\Phi(r)p(t,z) ,q_{\Phi,\eta}(r|t,z)p(t,z))$. In practice, we transform $q_{\Phi,\eta}(r|t,z)$ to $w_\eta(t,z,\textbf{\emph{x}})p_\Phi(r|t,z)$ by applying the change of variables formula twice as, 
\begin{equation}
    \begin{aligned}
        q_{\Phi,\eta}(r|t,z) &= q_{\Phi,\eta}(\Psi(r)|t,z)\cdot|det(J_{\Psi})|\\
        & = w_\eta(t,z, \Psi(r))p_\Phi(\Psi(r)|t,z)\cdot|det(J_{\Psi})|\\
        & = w_\eta(t,z, \Psi(r))p_\Phi(r|t,z). \label{equ:ipm}
    \end{aligned}    
\end{equation}
Then the IPM term becomes $IPM( p_\Phi(r)p(t,z) ,w_\eta(t,z,\textbf{\emph{x}})p_\Phi(r,t,z))$
and we use the Wasserstein distance as IPM.
To calculate it, we simulate samples with distribution $p_\Phi(r)p(t,z)$ by randomly permuting the observed treatment pairs, and the original samples are drawn from $p_\Phi(r,t,z)$, so the Wasserstein distance can be computed with $w_\eta(t,z,\textbf{\emph{x}})$.

\textbf{Estimation Module}. To predict the potential outcomes, we simply use an MLP as the predictor, which takes representation $r$, individual treatment $t$ and neighborhood exposure $z$ as inputs. The loss function of this module is
\begin{small}
    \begin{equation}
    \begin{aligned}
    \epsilon _{F}^{\eta}
    &=E_{p(t,z)}\left[\int _{\textbf{\emph{x}}} \ell _{h,\Phi }( \textbf{\emph{x}},t,z) q^{\eta}( \textbf{\emph{x}}|t,z) \ d\textbf{\emph{x}} \right]\\
    &=E_{p(t,z)}\left[\int _{\textbf{\emph{x}}} \ell _{h,\Phi }( \textbf{\emph{x}},t,z) w_\eta(t,z,\textbf{\emph{x}})p(\textbf{\emph{x}}|t,z) \ d\textbf{\emph{x}} \right]\\
    &=E_{p(\textbf{\emph{x}},t,z)}\left[w_\eta(t,z,\textbf{\emph{x}}) \ell _{h,\Phi }( \textbf{\emph{x}},t,z) \right].
    \end{aligned}
\end{equation}
\end{small}

Finally, the loss function induced by Theorem 
\ref{theo:ite bound} consists of two terms: the reweighted factual loss and the Wasserstein distance:
\begin{small}
    \begin{equation}
    \begin{aligned}
   L_{\mathcal{Y}}&=\frac{1}{n}\sum_{i=1}^nw_\eta(t_i,z_i,\textbf{\emph{x}}_i)(y_i-h_\Phi(r_i,t_i,z_i))^2\\
   &+\lambda \cdot \mathcal{W}( p_\Phi(r)p(t,z) ,w_\eta(t,z, \textbf{\emph{x}})p_\Phi(r,t,z)), \label{equ:finalLoss}
\end{aligned}    
\end{equation}
\end{small}
where $\lambda$ is the strength of the Wasserstein distance.

\textbf{Optimization}. Our entire model is optimized iteratively. At each iteration, we first train the individual propensity score estimator $(g_1, e)$ and neighborhood propensity score estimator $(g_2, \phi)$ by minimizing $L_{\mathcal{T}}$ and $L_{\mathcal{Z}}$, and then updates the GCN $m$, representation map $\Phi$ and predictor $h$ by minimizing $L_{\mathcal{Y}}$.

\section{EXPERIMENTS}
In this section, we validate the proposed method on two semi-synthetic datasets. In detail, we verify the effectiveness of our algorithm and further evaluate the correctness of the theorem analysis with the help of the semi-synthetic datasets.

\subsection{Datasets}
It is extremely challenging to collect the ground truth of ITEs because each unit can only be observed with one of the potential outcomes. Thus, we follow previous work \cite{jiang2022estimating, guo2020learning, ma2021deconfounding, veitch2019using} to create semi-synthetic datasets, i.e., the networks (features, topology) are real but treatments and potential outcomes are simulated. We produce the semi-synthetic datasets from the following two datasets.

\textbf{BlogCatalog (BC)} is a social network with blog service. Each instance is a blogger, and each edge signifies the friendship between two bloggers. The features are the keywords of each blogger’s articles. \textbf{Flickr} is an image and video sharing service. Each instance is a user and each edge represents the social relationship between users. The features of each user represent a list of tags of interest. 
Following existing works, we use the linear discriminant analysis (LDA) technique \cite{blei2003latent} to reduce the features' dimensions to 10. Then we generate the treatments and potential outcomes according to the following rules.

\textbf{Treatments simulation}. For network data, the individual treatment  $t_i$ is not only affected by $i$'s features $x_i$, but also affected by its neighbors' features $\{x_j\}_{j\in \mathcal{N}_i}$. To satisfy this setting, we first generate unit $i$'s propensity score $p(t_i|x_i,\{x_j\}_{j\in \mathcal{N}_i})$, then generate $i$'s treatment according to the mean propensity score as follows:
\begin{equation*}
\begin{small}
\begin{aligned}
    ps_i=\sigma(\beta_1\cdot (w_1&x_i)+\beta_2\cdot(\frac{1}{|\mathcal{N}_i|}\sum_{j\in \mathcal{N}_i}w_2x_j)),  \quad
    t_i=
    \begin{cases}
    1& \text{if $ps_i>\overline{ps}$},\\
    0& \text{else},
    \end{cases}
\end{aligned}
\end{small}
\end{equation*}
where $\sigma(\cdot )$ is sigmoid function, $\beta_1,\beta_2$ are strengths of individual features and neighbors' features, $w_1,w_2$ are weights generated by $\mathcal{N}(-2,3)$, and $\overline{ps}$ is the average of all $ps_i$. We set $\beta_1=0.5$ and set $\beta_2=k\beta_1$, where the control parameter $k\in\{0.5,1.0,2.0\}$ represents different degrees of neighbor interference. After generate individual treatment $t_i$,  the neighborhood exposure $z_i$, i.e.,the ratio of the treated neighbors of unit $i$, can be easily calculated by $z_i=\frac{1}{|\mathcal{N}_i|}\sum_{j\in \mathcal{N}_i}t_i$.

\textbf{Potential outcomes simulation}. The potential outcome $y_i(t_i,z_i)$ is affected by four factors: individual treatment $t_i$, neighborhood exposure $z_i$, $i$'s features $x_i$, $i$'s neighbors' features $\{x_j\}_{j\in \mathcal{N}_i}$. We generate potential outcome as follow:
\begin{equation*}
\begin{small}
\begin{aligned}
    y_i(t_i,z_i)=\beta_3\cdot t_i+\beta_4\cdot z_i+\beta_5\cdot(w_3x_i)+\beta_6\cdot(\frac{1}{|\mathcal{N}_i|}\sum_{j\in \mathcal{N}_i}w_4x_j)+\epsilon_i,
\end{aligned}
\end{small}
\end{equation*}
where $\epsilon_i$ is the noise sampled from $\mathcal{N}(0,1)$, $w_3,w_4$ are weights generated by $\mathcal{N}(1,2)$. The parameters $\beta_3,\beta_4,\beta_5,\beta_6$ are strengths of individual treatment, neighborhood exposure, individual features, and neighbors' features, respectively. We set $\beta_3=1,\;\beta_5=0.5$, and like the treatment simulation, we set $\beta_4=k\beta_3$, $\beta_6=k\beta_5$, where control parameter $k\in\{0.5,1.0,2.0\}$ represents different degrees of neighbor interference.

\begin{figure*}[htbp]
    \centering
    \vspace{-1mm}
    \setlength{\abovecaptionskip}{0.cm}
    \includegraphics[width=0.256\linewidth]{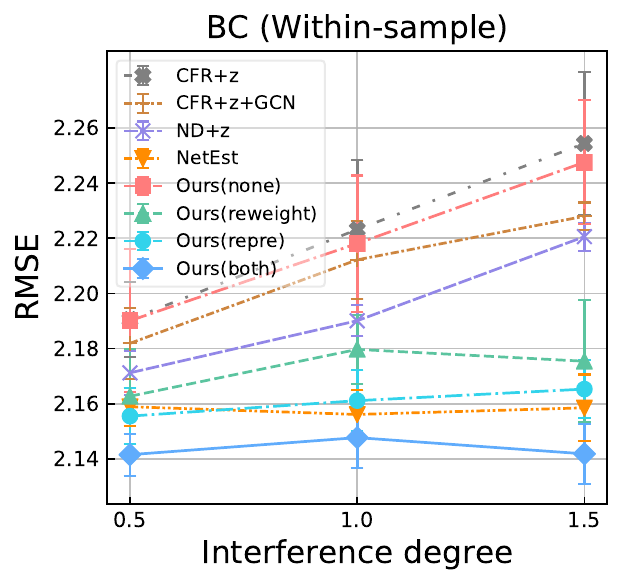}
    \includegraphics[width=0.24\linewidth]{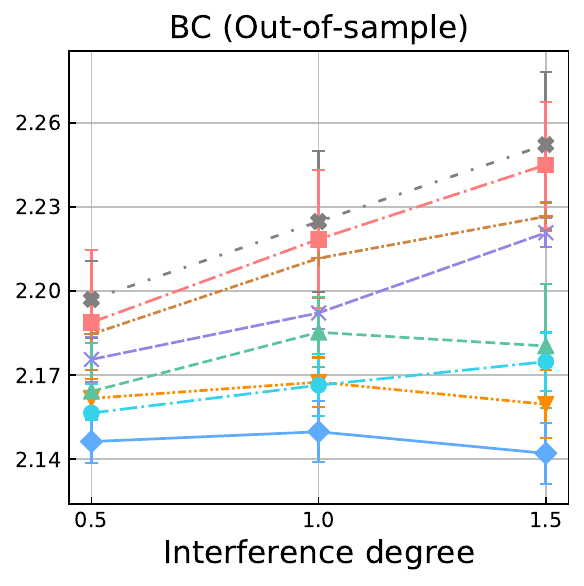}
    \includegraphics[width=0.24\linewidth]{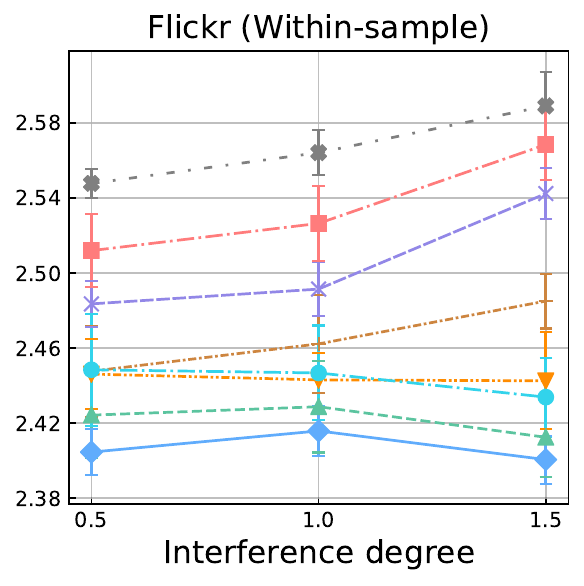}
    \includegraphics[width=0.24\linewidth]{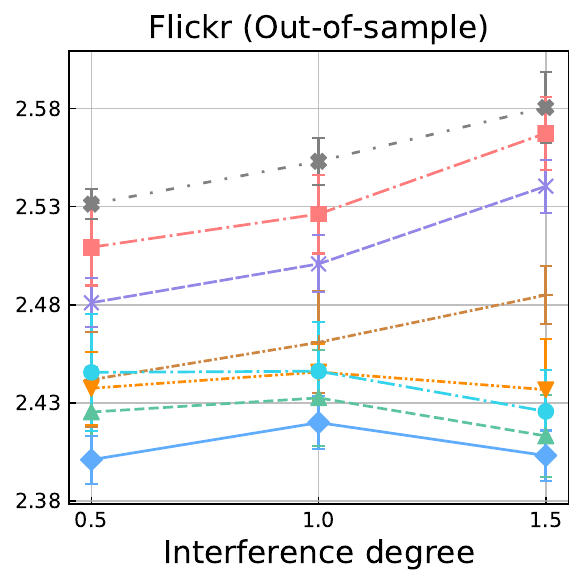}
    \caption{Counterfactual estimation errors $\epsilon_{RMSE} $ under different network interference degrees. From left: BlogCatalog “within-sample”, BlogCatalog “out-of-sample”, Flickr “within-sample”, Flickr “out-of-sample”.}
    \label{fig:rmse}
\end{figure*}

\begin{table*}[]
\setlength{\abovecaptionskip}{0cm}
\caption{Results of individual causal effect estimation on BC Dataset. The PEHE error $\epsilon_{PEHE}$(precision of estimating heterogeneous effects) is reported. The best-performing method is bolded.}
\label{table:bc}
\resizebox{\textwidth}{!}{
\begin{tabular}{ccccccccccc}
\toprule[1.5pt]
\makecell[c]{interference\\degree}                       & setting                        & effect     & CFR+z         & CFR+z+GCN     & ND+z          & NetEst        & Ours(none)    & Ours(repre)   & Ours(reweight) & Ours(both)    \\ \hline
\multicolumn{1}{c|}{\multirow{6}{*}{0.5}} & \multirow{3}{*}{Within Sample} & main & $0.8190_{\pm0.0352}$ & $0.8093_{\pm0.0654}$ & $0.8130_{\pm0.0677}$ & $0.6268_{\pm0.0717}$ & $0.8042_{\pm0.0911}$ & $0.5562_{\pm0.0221}$ & $0.6086_{\pm0.0175}$ & \pmb{$0.5410_{\pm0.0496}$} \\
\multicolumn{1}{c|}{}                     &                                & spillover       & $0.5777_{\pm0.0501}$ & $0.4772_{\pm0.0372}$ & $0.4385_{\pm0.0267}$ & $0.2910_{\pm0.0255}$ & $0.5340_{\pm0.0101}$ & \pmb{$0.2515_{\pm0.0426}$} & $0.4281_{\pm0.0449}$ & $0.2564_{\pm0.0324}$ \\
\multicolumn{1}{c|}{}                     &                                & total      & $1.3019_{\pm0.0419}$ & $0.9396_{\pm0.0943}$ & $0.8341_{\pm0.0612}$ & $0.5933_{\pm0.0654}$ & $1.1605_{\pm0.0306}$ & $0.5936_{\pm0.0475}$ & $0.9246_{\pm0.0576}$ & \pmb{$0.5609_{\pm0.0625}$} \\ \cline{2-11} 
\multicolumn{1}{c|}{}                     & \multirow{3}{*}{Out-of Sample} & main & $0.8192_{\pm0.0350}$ & $0.8588_{\pm0.0606}$ & $0.8119_{\pm0.0671}$ & $0.6126_{\pm0.0796}$ & $0.8089_{\pm0.0226}$ & $0.5642_{\pm0.0251}$ & $0.5936_{\pm0.0321}$ & \pmb{$0.5407_{\pm0.0497}$} \\
\multicolumn{1}{c|}{}                     &                                & spillover       & $0.5776_{\pm0.0512}$ & $0.4933_{\pm0.0252}$ & $0.4390_{\pm0.0262}$ & $0.3142_{\pm0.0528}$ & $0.5366_{\pm0.0263}$ & \pmb{$0.2543_{\pm0.0428}$} & $0.4248_{\pm0.0548}$ & $0.2663_{\pm0.0322}$ \\
\multicolumn{1}{c|}{}                     &                                & total      & $1.3017_{\pm0.0421}$ & $0.9244_{\pm0.0204}$ & $0.8437_{\pm0.0507}$ & $0.5962_{\pm0.0660}$ & $1.1594_{\pm0.0429}$ & $0.5918_{\pm0.0472}$ & $0.9161_{\pm0.0765}$ & \pmb{$0.5712_{\pm0.0630}$} \\ \hline

\multicolumn{1}{c|}{\multirow{6}{*}{1}}   & \multirow{3}{*}{Within Sample} & main & $0.8952_{\pm0.0856}$ & $0.8228_{\pm0.0450}$ & $0.7436_{\pm0.0665}$ & $0.6676_{\pm0.0228}$ & $0.8583_{\pm0.0138}$ & $0.6502_{\pm0.0630}$ & $0.6585_{\pm0.0778}$ & \pmb{$0.6189_{\pm0.0649}$} \\
\multicolumn{1}{c|}{}                     &                                & spillover       & $1.0054_{\pm0.0514}$ & $0.9355_{\pm0.0404}$ & $0.8823_{\pm0.0405}$ & $0.3612_{\pm0.0386}$ & $0.9624_{\pm0.0176}$ & $0.3882_{\pm0.0637}$ & $0.4134_{\pm0.0613}$ & \pmb{$0.2537_{\pm0.0401}$} \\
\multicolumn{1}{c|}{}                     &                                & total      & $2.3215_{\pm0.0816}$ & $1.5786_{\pm0.0187}$ & $1.1277_{\pm0.0457}$ & $0.5279_{\pm0.0807}$ & $1.8724_{\pm0.0177}$ & $0.5548_{\pm0.0481}$ & $0.9796_{\pm0.0862}$ & \pmb{$0.5036_{\pm0.0554}$} \\ \cline{2-11} 
\multicolumn{1}{c|}{}                     & \multirow{3}{*}{Out-of Sample} & main & $0.8951_{\pm0.0845}$ & $0.8160_{\pm0.0552}$ & $0.7420_{\pm0.0655}$ & \pmb{$0.6333_{\pm0.0412}$} & $0.8463_{\pm0.0141}$ & $0.6506_{\pm0.0628}$ & $0.6459_{\pm0.0704}$ & $0.6388_{\pm0.0650}$ \\
\multicolumn{1}{c|}{}                     &                                & spillover       & $1.0043_{\pm0.0525}$ & $0.9409_{\pm0.0432}$ & $0.8835_{\pm0.0387}$ & $0.3832_{\pm0.0377}$ & $0.9695_{\pm0.0159}$ & $0.3858_{\pm0.0613}$ & $0.4234_{\pm0.0607}$ & \pmb{$0.2538_{\pm0.0403}$} \\
\multicolumn{1}{c|}{}                     &                                & total      & $2.3213_{\pm0.0818}$ & $1.5674_{\pm0.0199}$ & $1.1328_{\pm0.0440}$ & $0.5313_{\pm0.0908}$ & $1.8824_{\pm0.0258}$ & $0.5546_{\pm0.0282}$ & $0.9696_{\pm0.0810}$ & \pmb{$0.5036_{\pm0.0555}$} \\ \hline

\multicolumn{1}{c|}{\multirow{6}{*}{1.5}} & \multirow{3}{*}{Within Sample} & main & $0.9304_{\pm0.0490}$ & $0.8409_{\pm0.0512}$ & $0.8023_{\pm0.0616}$ & $0.6101_{\pm0.0318}$ & $0.9431_{\pm0.0618}$ & $0.6237_{\pm0.0250}$ & $0.6190_{\pm0.0836}$ & \pmb{$0.6048_{\pm0.0108}$} \\
\multicolumn{1}{c|}{}                     &                                & spillover       & $1.4989_{\pm0.0542}$ & $1.4373_{\pm0.0627}$ & $1.3404_{\pm0.0677}$ & $0.3804_{\pm0.0370}$ & $1.2698_{\pm0.0748}$ & $0.3894_{\pm0.0827}$ & $0.4341_{\pm0.0980}$ & \pmb{$0.3613_{\pm0.0386}$} \\
\multicolumn{1}{c|}{}                     &                                & total      & $3.3342_{\pm0.0671}$ & $2.2433_{\pm0.0485}$ & $2.2573_{\pm0.0286}$ & \pmb{$0.5469_{\pm0.0951}$} & $3.0156_{\pm0.0746}$ & $0.5715_{\pm0.0860}$ & $0.9875_{\pm0.0214}$ & $0.5863_{\pm0.0310}$ \\ \cline{2-11} 
\multicolumn{1}{c|}{}                     & \multirow{3}{*}{Out-of Sample} & main & $0.9303_{\pm0.0492}$ & $0.8604_{\pm0.0711}$ & $0.8119_{\pm0.0780}$ & \pmb{$0.6001_{\pm0.0514}$} & $0.9554_{\pm0.0848}$ & $0.6322_{\pm0.0236}$ & $0.6219_{\pm0.0868}$ & $0.6101_{\pm0.0128}$ \\
\multicolumn{1}{c|}{}                     &                                & spillover       & $1.4985_{\pm0.0585}$ & $1.4395_{\pm0.0678}$ & $1.3441_{\pm0.0653}$ & $0.3880_{\pm0.0414}$ & $1.2795_{\pm0.0701}$ & $0.4010_{\pm0.0948}$ & $0.4429_{\pm0.0878}$ & \pmb{$0.3691_{\pm0.0453}$} \\
\multicolumn{1}{c|}{}                     &                                & total      & $3.3341_{\pm0.0672}$ & $2.2702_{\pm0.0651}$ & $2.2633_{\pm0.0439}$ & \pmb{$0.5358_{\pm0.0830}$} & $3.0135_{\pm0.0895}$ & $0.5896_{\pm0.0880}$ & $0.9989_{\pm0.0131}$ & $0.5843_{\pm0.0320}$ \\
\bottomrule[1.5pt]
\end{tabular}
}
\end{table*}

\subsection{Baselines}
We compare the performance of our algorithm with four baselines and three variants.
\textbf{CFR+$z_i$} \cite{shalit2017estimating}: CFR is an ITE estimator for i.i.d data. We add the neighborhood exposure $z_i$ as extra input to evaluate its performance under network interference.
\textbf{CFR+$z_i$+GCN} \cite{ma2021causal}: Based on CFR, this model applies GCN to aggregate neighbors' features, then adds it together with $z_i$ as extra input.
\textbf{ND+$z_i$} \cite{guo2020learning}: ND learns latent confounders' representations using GCN and minimizes the IPM to obtain a balanced representation.
\textbf{NetEst} \cite{jiang2022estimating}: State-of-the-art model for estimating ITE from observational network data. It bridges the distribution gaps between the graph ML model and causal effect estimation via adversarial learning.
\textbf{Ours(none), Ours(reweight), Ours(repre)}: Variants of our algorithm without reweighting and representation learning, only with reweighting and only with representation learning, respectively.

\textbf{Implementation details}. We build our model as follows. We use one graph convolution layer and set its embedding size as 10. For the sparse encoder $g_1(\cdot), g_2(\cdot)$, representation map $\Phi$ and predictor $h$, we use three fully-connected layers and set all hidden sizes as 32. For the decoder $e(\cdot)$ and $\phi(\cdot)$, we use logistic regression and piecewise linear functions, respectively.
For hyperparameters, we use full-batch training and use Adam optimizer with learning rate=0.001. 
For the parameters in loss functions, we fix $\beta=0.001$, $\rho=0.05$, and the space of parameter $\lambda$ is $\{0.1, 0.2, 0.5\}$. 

\subsection{Results}
We consider two metrics for performance evaluation. The first one is \textbf{Root Mean Squared Error} for counterfactual estimation, which is calculated as $\epsilon_{RMSE}=\sqrt{\frac{1}{N}\sum_{i=1}^{N}(\hat{y}_i^{CF}-y_i^{CF})^2}$. Note that $y_i^{CF}$ is the outcomes with flipped individual treatment $\tilde{t}_i$ (if $t_i=1,$ then $\tilde{t}_i=0$ and vice versa) and  corresponding $\tilde{z_i}=\frac{1}{|\mathcal{N}_i|}\sum_{j\in \mathcal{N}_i}\tilde{t}_i$. The other is \textbf{the Precision in Estimation of Heterogeneous Effect} $\epsilon_{PEHE}=\sqrt{\frac{1}{N}\sum_{i=1}^{N}(\hat{\tau}(x)-\tau(x))^2}$. Here we consider three types of causal effects: main effect, spillover effect, and total effect. In addition, we use METIS \cite{karypis1998fast} to partition the original network into three sub-networks: train, valid and test. Then we evaluate the "within-sample" estimation on train network and "out-of-sample" on the test network. We run every task five times and reported the average and standard deviation. The experimental result of $\epsilon_{RMSE}$ is shown in Fig. \ref{fig:rmse}, and the results of $\epsilon_{PEHE}$ on BC and Flickr datasets are reported in Table \ref{table:bc} and Table \ref{table:flickr}, respectively. 

In general, for both the counterfactual estimation task and causal effect estimation task, Ours(both) consistently outperforms all baselines in “within-sample” and “out-of-sample” settings under different neighbor interference degrees. First, it indicates our designed objective function indeed minimizes the errors of predicting counterfactual outcomes and ITEs, which verifies Lemma \ref{lem: cf bound} and Theorem \ref{theo:ite bound}. Second, it suggests our model is of effectiveness and robustness compared to the others. Specifically, from the results, we can also draw the following conclusions:

\begin{table*}[]
\setlength{\abovecaptionskip}{0cm}
\caption{Results of individual causal effect estimation on Flickr Dataset. The PEHE error $\epsilon_{PEHE}$(precision of estimating heterogeneous effects) is reported. The best performing method is bolded.} \label{table:flickr}
\resizebox{\textwidth}{!}{
    \begin{tabular}{ccccccccccc}
\toprule[1.5pt]
\makecell[c]{interference\\degree}                       & setting                        & effect     & CFR+z         & CFR+z+GCN     & ND+z          & NetEst        & Ours(none)    & Ours(repre)   & Ours(reweight) & Ours(both)    \\ \hline
\multicolumn{1}{c|}{\multirow{6}{*}{0.5}} & \multirow{3}{*}{Within Sample} & individual & $1.9630_{\pm0.0533}$ & \pmb{$0.9854_{\pm0.0700}$} & $1.7103_{\pm0.0648}$ & $1.5834_{\pm0.0661}$ & $1.8846_{\pm0.0394}$ & $1.6521_{\pm0.0292}$ & $1.1539_{\pm0.0597}$ & $1.0884_{\pm0.0198}$ \\
\multicolumn{1}{c|}{}                     &                                & peer       & $0.4870_{\pm0.0315}$ & $0.3967_{\pm0.0357}$ & $0.4082_{\pm0.0304}$ & $0.2103_{\pm0.0349}$ & $0.4688_{\pm0.0806}$ & $0.2414_{\pm0.0593}$ & $0.1919_{\pm0.0286}$  & \pmb{$0.1229_{\pm0.0622}$} \\
\multicolumn{1}{c|}{}                     &                                & total      & $2.6071_{\pm0.0552}$ & $1.5457_{\pm0.0598}$ & $2.3370_{\pm0.0166}$ & $1.8020_{\pm0.0842}$ & $2.5122_{\pm0.0473}$ & $1.8363_{\pm0.0898}$ & $1.5145_{\pm0.0607}$  & \pmb{$1.3016_{\pm0.0397}$} \\ \cline{2-11} 
\multicolumn{1}{c|}{}                     & \multirow{3}{*}{Out-of Sample} & individual & $1.9629_{\pm0.0526}$ & \pmb{$0.9716_{\pm0.0537}$} & $1.7108_{\pm0.0647}$ & $1.5813_{\pm0.0220}$ & $1.8672_{\pm0.0279}$ & $1.6505_{\pm0.0323}$ & $1.1383_{\pm0.0559}$  & $1.0748_{\pm0.0195}$ \\
\multicolumn{1}{c|}{}                     &                                & peer       & $0.4869_{\pm0.0342}$ & $0.3991_{\pm0.0359}$ & $0.4080_{\pm0.0303}$ & $0.2305_{\pm0.0397}$ & $0.4547_{\pm0.0955}$ & $0.2454_{\pm0.0593}$ & $0.1795_{\pm0.0163}$  & \pmb{$0.1189_{\pm0.0622}$} \\
\multicolumn{1}{c|}{}                     &                                & total      & $2.5971_{\pm0.0524}$ & $1.5232_{\pm0.0426}$ & $2.3375_{\pm0.0166}$ & $1.8213_{\pm0.0137}$ & $2.5071_{\pm0.0578}$ & $1.8351_{\pm0.0921}$ & $1.4918_{\pm0.0565}$  & \pmb{$1.2966_{\pm0.0461}$} \\ \hline

\multicolumn{1}{c|}{\multirow{6}{*}{1}}   & \multirow{3}{*}{Within Sample} & individual & $2.2184_{\pm0.0576}$ & $1.0984_{\pm0.0738}$ & $1.9813_{\pm0.0741}$ & $1.5404_{\pm0.0829}$ & $2.0783_{\pm0.0591}$ & $1.6249_{\pm0.0854}$ & $1.1237_{\pm0.0438}$  & \pmb{$1.0763_{\pm0.0635}$} \\
\multicolumn{1}{c|}{}                     &                                & peer       & $0.8726_{\pm0.0390}$ & $0.8069_{\pm0.0471}$ & $0.6870_{\pm0.0327}$ & $0.3376_{\pm0.0386}$ & $0.8369_{\pm0.0307}$ & $0.3837_{\pm0.0628}$ & $0.3552_{\pm0.0514}$  & \pmb{$0.2781_{\pm0.0528}$} \\
\multicolumn{1}{c|}{}                     &                                & total      & $3.5033_{\pm0.0769}$ & $1.9979_{\pm0.0940}$ & $3.0581_{\pm0.0258}$ & $1.8209_{\pm0.0291}$ & $3.2233_{\pm0.0649}$ & $1.9422_{\pm0.0996}$ & $1.8124_{\pm0.0589}$  & \pmb{$1.3592_{\pm0.0779}$} \\ \cline{2-11} 
\multicolumn{1}{c|}{}                     & \multirow{3}{*}{Out-of Sample} & individual & $2.2185_{\pm0.0524}$ & $1.0823_{\pm0.0611}$ & $2.0063_{\pm0.0295}$ & $1.5369_{\pm0.0134}$ & $2.0738_{\pm0.0454}$ & $1.6239_{\pm0.0867}$ & $1.0984_{\pm0.0389}$  & \pmb{$1.0667_{\pm0.0663}$} \\
\multicolumn{1}{c|}{}                     &                                & peer       & $0.8724_{\pm0.0325}$ & $0.8086_{\pm0.0494}$ & $0.6764_{\pm0.0370}$ & $0.3412_{\pm0.0614}$ & $0.8451_{\pm0.0402}$ & $0.3717_{\pm0.0628}$ & $0.3444_{\pm0.0452}$  & \pmb{$0.2767_{\pm0.0524}$} \\
\multicolumn{1}{c|}{}                     &                                & total      & $3.5213_{\pm0.0726}$ & $1.9771_{\pm0.0908}$ & $3.0782_{\pm0.0538}$ & $1.8438_{\pm0.0540}$ & $3.2175_{\pm0.0676}$ & $1.9407_{\pm0.0336}$ & $1.7933_{\pm0.0462}$  & \pmb{$1.3626_{\pm0.0847}$} \\ \hline

\multicolumn{1}{c|}{\multirow{6}{*}{1.5}} & \multirow{3}{*}{Within Sample} & individual & $2.3601_{\pm0.0589}$ & $1.0470_{\pm0.0696}$ & $2.0884_{\pm0.0343}$ & $1.5040_{\pm0.0231}$ & $2.2359_{\pm0.0766}$ & $1.6228_{\pm0.0390}$ & \pmb{$0.9895_{\pm0.0210}$}  & $1.0195_{\pm0.0622}$ \\
\multicolumn{1}{c|}{}                     &                                & peer       & $1.4488_{\pm0.0433}$ & $1.2681_{\pm0.0536}$ & $1.2873_{\pm0.0323}$ & $0.3798_{\pm0.0435}$ & $1.3662_{\pm0.0817}$ & $0.3629_{\pm0.0653}$ & $0.3181_{\pm0.0238}$  & \pmb{$0.2554_{\pm0.0487}$} \\
\multicolumn{1}{c|}{}                     &                                & total      & $4.3957_{\pm0.1030}$ & $2.6519_{\pm0.0969}$ & $3.8458_{\pm0.0555}$ & $1.8563_{\pm0.0949}$ & $3.9306_{\pm0.0923}$ & $1.8283_{\pm0.0628}$ & $1.7622_{\pm0.0883}$  & \pmb{$1.3937_{\pm0.0127}$} \\ \cline{2-11} 
\multicolumn{1}{c|}{}                     & \multirow{3}{*}{Out-of Sample} & individual & $2.3824_{\pm0.0526}$ & $1.0497_{\pm0.0683}$ & $2.0855_{\pm0.0129}$ & $1.4871_{\pm0.0375}$ & $2.2252_{\pm0.0599}$ & $1.6228_{\pm0.0130}$ & \pmb{$1.0039_{\pm0.0134}$}  & $1.0059_{\pm0.0671}$ \\
\multicolumn{1}{c|}{}                     &                                & peer       & $1.4445_{\pm0.0457}$ & $1.2750_{\pm0.0507}$ & $1.2882_{\pm0.0320}$ & $0.3854_{\pm0.0704}$ & $1.3432_{\pm0.0860}$ & $0.3409_{\pm0.0624}$ & $0.3100_{\pm0.0872}$  & \pmb{$0.2548_{\pm0.0487}$} \\
\multicolumn{1}{c|}{}                     &                                & total      & $4.3847_{\pm0.0525}$ & $2.6625_{\pm0.0999}$ & $3.8576_{\pm0.0627}$ & $1.8612_{\pm0.0226}$ & $3.9421_{\pm0.0942}$ & $1.8262_{\pm0.0666}$ & $1.7983_{\pm0.0685}$  & \pmb{$1.3978_{\pm0.0203}$} \\ 
\bottomrule[1.5pt]
\end{tabular}
}
\vspace{-1mm}
\end{table*}

\textbf{It is not enough to simply add additional input variables to the existing models}. 
As shown in Fig. \ref{fig:rmse}, as the neighbor interference degree increases, the performances of our model and NetEst keep stable while the performances of CFR+$z_i$, CFR+$z_i$+GCN and ND+$z_i$ become worse significantly. The reason is that CFR+$z_i$, CFR+$z_i$+GCN and ND+$z_i$ simply treat the neighbors' features or neighborhood exposure as additional input variables, which neglects the complex confounding bias in network data. In contrast, our model and NetEst seek to address the complex confounding bias in network data. Therefore, we can conclude that the complex confounding bias in network data should be considered more carefully instead of simply treating the neighbors' features or neighborhood exposure as additional inputs.

\textbf{Single reweighting or representation learning methods can alleviate confounding bias to some extent}. As shown in Table \ref{table:bc} and \ref{table:flickr}, Ours(reweight) and Ours(repre) outperform Ours(none) on both tasks, indicating that either reweighting or representation learning methods are effective to reduce the confounding bias. 
In addition, NetEst is essentially a representation learning- based method, which uses the adversarial loss to replace the IPM term in Ours(repre), and thus NetEst has a similar performance to Ours(repre), which also means the representation learning is able to alleviate confounding bias.

\begin{figure}[!h]
    \centering
    \vspace{-1mm}
    \setlength{\abovecaptionskip}{0.cm}
    \includegraphics[width=0.46\linewidth]{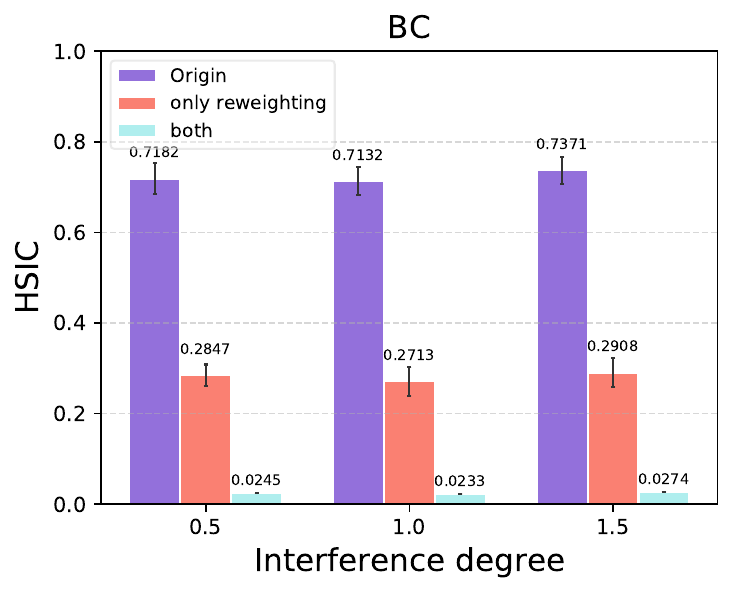} \quad
    \includegraphics[width=0.45\linewidth]{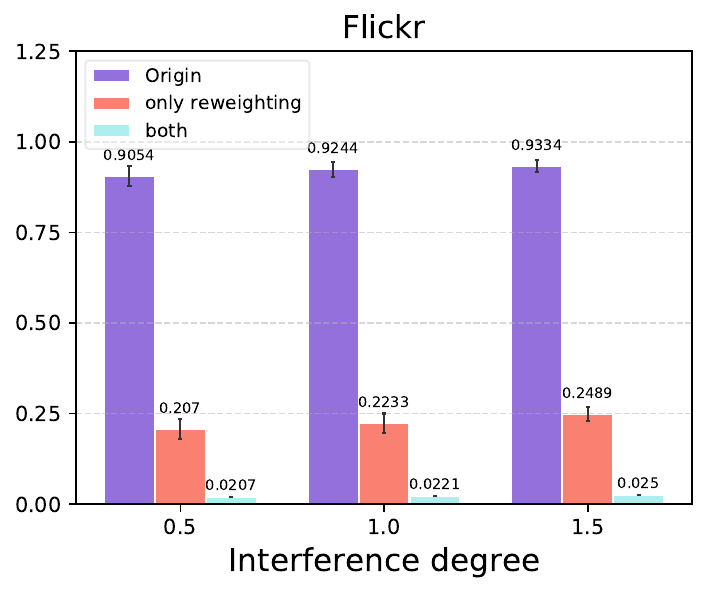}
    \caption{HSIC between feature distribution $p(\cdot)$ and treatment pair distribution $p(t,z)$ after (1) no any operation (purple); (2) reweighting (red); (3)  reweighting enhanced with representation learning (light blue).}
    \label{fig:wass}
    \vspace{-2mm}
\end{figure}

\textbf{The combination of reweighting and representation learning can enhance each other}. As shown in Table \ref{table:bc} and \ref{table:flickr}, Ours(both) outperforms Ours(repre) and Ours(reweight), indicating that the reweighting and representation learning can enhance each other and further improve performance:
\begin{itemize}[leftmargin = 10 pt, topsep = 2 pt]
    \item Compared to Ours(repre), Ours(both) incorporates balanced weights to get a tighter bound, which can lead to better performance and also verifies Theorem \ref{theo:tighter bound}. 
    \item  Compared to Ours(reweight), Ours(both) addresses confounding bias
    manifested by imperfections in the estimated joint propensity score by forcing the reweighted representation balanced. We check the conclusion by additionally comparing the dependence between the treatment pair and reweighted features of these two methods using $HSIC$. As shown in Fig \ref{fig:wass}, the dependence between the reweighted features and treatment pair (red bar) is significantly reduced compared with original features (purple bar). However, due to imperfect estimation of the joint propensity score, independence is still not achieved with Ours(reweight), while Ours(Both) achieves it by learning balanced representation from the reweighted features (light blue bar). 
\end{itemize}

In addition, Fig. \ref{fig:wass} also explains why Ours(reweight) performs better than Ours(repre) on Flickr in Table \ref{table:flickr}, and the opposite is true on BC in Table \ref{table:bc}: after reweighting, the dependence between origin features and treatment pair is reduced more on Flickr (75.43\% reduction on average) than on BC (61.34\% reduction on average), probably because the joint propensity score is estimated more accurately on Flickr.

\section{CONCLUSION}
In this paper, we study the problem of estimating ITE from observational network data. 
Specifically, we derive the generalization bound by exploiting the reweighting and representation learning simultaneously. We theoretically show that introducing representation learning is able to address the bias caused by the inaccurate estimated joint propensity scores in the reweighting approach, and introducing balancing weights can make the generalization bound tighter compared to purely representation learning. Based on the above theoretical results, we finally propose a weighting regression method utilizing the joint propensity score augmented with representation learning. The effectiveness of our method is demonstrated on two semi-synthetic datasets.

\begin{acks}
To Robert, for the bagels and explaining CMYK and color spaces.
\end{acks}

\appendix

\section{Proof of Lemma \ref{lem: cf bound}}
For simplicity, we will use $\textbf{\emph{t}}=(t,z)$ to denote the vector consisting of individual treatment $t$ and neighborhood exposure $z$. 
\begin{proof}
We first simplify $\epsilon _{CF}(\textbf{\emph{t}})$:
\begin{small}
    \begin{equation}
    \begin{aligned}
        \epsilon _{CF}(\textbf{\emph{t}}) & =E_{p( \textbf{\emph{t}}^{\prime})}\int _{\mathcal{X}} l_{h,\phi }( \textbf{\emph{x}},\textbf{\emph{t}}) p( \textbf{\emph{x}}|\textbf{\emph{t}}^{\prime}) \ d\textbf{\emph{x}}\\
         & =\int _{\mathcal{T}^{\prime}}\int _{\mathcal{X}} l_{h,\phi }( \textbf{\emph{x}},\textbf{\emph{t}}) p( \textbf{\emph{x}}|\textbf{\emph{t}}^{\prime}) \ p( \textbf{\emph{t}}^{\prime}) d\textbf{\emph{x}}d\textbf{\emph{t}}^{\prime}\\
         & =\int _{\mathcal{X}} l_{h,\phi }( \textbf{\emph{x}},\textbf{\emph{t}}) p( \textbf{\emph{x}}) d\textbf{\emph{x}}. \nonumber
    \end{aligned}
\end{equation}
\end{small}

Then we have:
\begin{small}
    \begin{equation}
    \begin{aligned}
         \epsilon _{CF} -\epsilon _{F}^\eta 
        = & \int _{\mathcal{T}}\int _{\mathcal{X}} l_{h,\phi }( \textbf{\emph{x}},\textbf{\emph{t}}) p( \textbf{\emph{x}}) p(\textbf{\emph{t}}) \ d\textbf{\emph{x}}d\textbf{\emph{t}} \\
        & -\int _{\mathcal{T}}\int _{\mathcal{X}} l_{h,\phi }( \textbf{\emph{x}},\textbf{\emph{t}}) q_\eta( \textbf{\emph{x}}|\textbf{\emph{t}})p(\textbf{\emph{t}}) \ d\textbf{\emph{x}}d\textbf{\emph{t}}\\
         =&B_{\Phi}\int _{\mathcal{T}}\int _{\mathcal{X}} \frac{1}{B_{\Phi}}\cdot l_{h,\phi }( x,\textbf{\emph{t}})( p( \textbf{\emph{x}}) p(\textbf{\emph{t}}) -q_\eta( \textbf{\emph{x}}|\textbf{\emph{t}})p(\textbf{\emph{t}})) \ d\textbf{\emph{x}}d\textbf{\emph{t}}\\
         \leqslant & B_{\Phi}\cdot \sup _{g\in G} |\int _{\mathcal{T}}\int _{\mathcal{X}} g( \textbf{\emph{x}},\textbf{\emph{t}}) \cdot ( p( \textbf{\emph{x}}) p(\textbf{\emph{t}}) -q_\eta( \textbf{\emph{x}}|\textbf{\emph{t}})p(\textbf{\emph{t}})) \ d\textbf{\emph{x}}d\textbf{\emph{t}} \\
        =&B_{\Phi}\cdot IPM(p( \textbf{\emph{x}}) p(\textbf{\emph{t}}),q_\eta( \textbf{\emph{x}}|\textbf{\emph{t}})p(\textbf{\emph{t}})), \nonumber
    \end{aligned}
\end{equation}
\end{small}
where the third formula is from the definition of IPM. Then after applying a change of variable $r$ to the space induced by the representation $\Phi$, we obtain Lemma \ref{lem: cf bound}. 
\end{proof}

\section{Proof of Theorem \ref{theo:ite bound}}
We define $\hat{\tau }_{h,\Phi}^{\textbf{\emph{t}} ,\textbf{\emph{t}}^{\prime}}( \textbf{\emph{x}})=h(\Phi(\textbf{\emph{x}}),\textbf{\emph{t}} ) -h(\Phi(\textbf{\emph{x}}), \textbf{\emph{t}}^{\prime})$ and $\tau ^{\textbf{\emph{t}} ,\textbf{\emph{t}}^{\prime}}( \textbf{\emph{x}})=m(\textbf{\emph{x}},\textbf{\emph{t}} ) -m( \textbf{\emph{x}},\textbf{\emph{t}}^{\prime})$, where $m(\textbf{\emph{x}},\textbf{\emph{t}})=\int_{\mathcal{Y}}y(\textbf{\emph{t}})p( y(\textbf{\emph{t}}) |\textbf{\emph{x}})dy(\textbf{\emph{t}})$ and  $\textbf{\emph{t}}=(t,z)$.
\begin{proof}
\begin{small}
    \begin{equation*}
    \begin{aligned}
        &\epsilon_{PEHE}(h,\Phi) \\
        =& E_{p(\textbf{\emph{t}})p(\textbf{\emph{t}}^{\prime})}\left[\int _{\mathcal{X}}(\hat{\tau }_{h,\Phi}^{\textbf{\emph{t}} ,\textbf{\emph{t}}^{\prime}}( \textbf{\emph{x}}) -\tau ^{\textbf{\emph{t}} ,\textbf{\emph{t}}^{\prime}}( \textbf{\emph{x}}))^{2} p( \textbf{\emph{x}}) \ d\textbf{\emph{x}}\right] \\
        =&E_{p(\textbf{\emph{t}})p(\textbf{\emph{t}}^{\prime})}\left[\int _{\mathcal{X}}(h(\Phi(\textbf{\emph{x}}),\textbf{\emph{t}} ) -h(\Phi(\textbf{\emph{x}}), \textbf{\emph{t}}^{\prime}) -( m(\textbf{\emph{x}},\textbf{\emph{t}} ) -m( \textbf{\emph{x}},\textbf{\emph{t}}^{\prime})))^{2} p( \textbf{\emph{x}}) \ d\textbf{\emph{x}}\right]\\
        \leqslant& 2\int _{\mathcal{T}}\int _{\mathcal{X}}(( h(\Phi(\textbf{\emph{x}}),\textbf{\emph{t}}) -m(\textbf{\emph{x}},\textbf{\emph{t}} ))^{2} p( \textbf{\emph{x}}) p(\textbf{\emph{t}}) \ d\textbf{\emph{x}}d\textbf{\emph{t}}\\
         &+2\int _{\mathcal{T}^{\prime}}\int _{\mathcal{X}}(( h(\Phi(\textbf{\emph{x}}),\textbf{\emph{t}}^{\prime} ) -m(\textbf{\emph{x}}, \textbf{\emph{t}}^{\prime}))^{2} p( \textbf{\emph{x}}) p(\textbf{\emph{t}}^{\prime}) \ d\textbf{\emph{x}}d\textbf{\emph{t}}^{\prime}\\
        =&2( \epsilon _{CF} -\sigma _{Y}) +2( \epsilon _{CF} -\sigma _{Y})\\
         \leqslant& 4\left( \epsilon _{F}^\eta +B_{\Phi } \cdot IPM(p_\Phi( r) p(\textbf{\emph{t}}),q_{\Phi,\eta}(r|\textbf{\emph{t}})p(\textbf{\emph{t}}))-\sigma _{Y}\right),
    \end{aligned}
\end{equation*}
\end{small}
where the third formula holds since $(x+y)^2\leq2(x^2+y^2)$, the last formula holds by applying Lemma 1, and the third formula can be derived from the following derivation:
    \begin{small}
        \begin{equation}
            \begin{aligned}
            &\epsilon _{CF} \\
            =&\int _{\mathcal{T}}\int _{\mathcal{X}} l_{h,\phi }( \textbf{\emph{x}},\textbf{\emph{t}}) p( \textbf{\emph{x}}) p(\textbf{\emph{t}}) \ d\textbf{\emph{x}}d\textbf{\emph{t}}\\
            =&\int _{\mathcal{T}}\int _{\mathcal{X}}\int _{\mathcal{Y}}( y(\textbf{\emph{t}}) -h( \Phi(\textbf{\emph{x}}),\textbf{\emph{t}}))^{2} p( y(\textbf{\emph{t}}) |\textbf{\emph{x}}) p( \textbf{\emph{x}}) p(\textbf{\emph{t}}) \ dy(\textbf{\emph{t}}) d\textbf{\emph{x}}d\textbf{\emph{t}}\\
            =&\int _{\mathcal{T}}\int _{\mathcal{X}}\int _{\mathcal{Y}}( h( \Phi(\textbf{\emph{x}}),\textbf{\emph{t}}) -m( \textbf{\emph{x}},\textbf{\emph{t}}))^{2} p( y(\textbf{\emph{t}}) |\textbf{\emph{x}}) p( \textbf{\emph{x}}) p(\textbf{\emph{t}}) \ dy(\textbf{\emph{t}}) d\textbf{\emph{x}}d\textbf{\emph{t}}\\
             & \ +\ \int _{\mathcal{T}}\int _{\mathcal{X}}\int _{\mathcal{Y}}( m( \textbf{\emph{x}},\textbf{\emph{t}}) -y(\textbf{\emph{t}}))^{2} p( y(\textbf{\emph{t}}) |\textbf{\emph{x}}) p( \textbf{\emph{x}}) p(\textbf{\emph{t}}) \ dy(\textbf{\emph{t}}) d\textbf{\emph{x}}d\textbf{\emph{t}}\\
             & \ +\ noise\;term\\  
             =&\int _{\mathcal{T}}\int _{\mathcal{X}}(( h( \Phi(\textbf{\emph{x}}),\textbf{\emph{t}}) -m( \textbf{\emph{x}},\textbf{\emph{t}}))^{2} p( \textbf{\emph{x}}) p(\textbf{\emph{t}}) \ d\textbf{\emph{x}}d\textbf{\emph{t}} +\sigma _{Y},\nonumber
            \end{aligned}
        \end{equation}
    \end{small}
where the third formula holds by applying standard bias-noise decomposition and the noise term evaluates to zero.
We define the variance of $y(\textbf{\emph{t}})$ with respect to the distribution $p(x)p(\textbf{\emph{t}})$ as $\sigma _Y$.

\end{proof}

\section{Proof of Theorem \ref{theo:wass bound}}
Following \cite{assaad2021counterfactual}, we derive $D_{KL}(p_\Phi(r)p(\textbf{\emph{t}}), q_{_\Phi,\eta}(r|\textbf{\emph{t}})p(\textbf{\emph{t}}))$ first, and then use it to bound the total variation distance, which will be used to bound the $\mathcal{W}(p_\Phi(r)p(\textbf{\emph{t}}), q_{_\Phi,\eta}(r|\textbf{\emph{t}})p(\textbf{\emph{t}}))$ finally. The definition of total variation distance is as follows:

\begin{definition}
The total variation distance (TVD) between distributions $p$ and $q$ on $\mathcal{R}$ is defined as:
\begin{small}
    \begin{align}
    \delta(p,q)=\frac{1}{2}\cdot \sup_{m:\| m \|_\infty\leq1}\left[ \int_{\mathcal{R}} m(r)(p(r)-q(r))dr\right]. \nonumber
\end{align}
\end{small}
\end{definition}
Next, we prove Theorem \ref{theo:wass bound}.
\begin{proof}
First, we re-write the reweighted distribution as:
\begin{small}
        \begin{equation}
    \begin{aligned}
    q_{\eta }( \textbf{\emph{x}}|\textbf{\emph{t}}) &\varpropto \frac{1}{\psi _{\eta }( \textbf{\emph{t}};\textbf{\emph{x}})} p( \textbf{\emph{x}}|\textbf{\emph{t}}) 
    =\frac{1}{\psi ( \textbf{\emph{t}};\textbf{\emph{x}})} p( \textbf{\emph{x}}|\textbf{\emph{t}})\frac{\psi ( \textbf{\emph{t}};\textbf{\emph{x}})}{\psi _{\eta }( \textbf{\emph{t}};\textbf{\emph{x}})} \\
    &=p( \textbf{\emph{x}})\frac{e( t;\textbf{\emph{x}}_{t}) \phi ( z;\textbf{\emph{x}}_{z})}{e_{\eta }( t;\textbf{\emph{x}}_{t}) \phi _{\eta }( z;\textbf{\emph{x}}_{z})}. \nonumber
    \end{aligned}
\end{equation}
\end{small}

Next, we compute the KL-divergence between $p(\textbf{\emph{x}})p(\textbf{\emph{t}})$ and $q_\eta(\textbf{\emph{x}}|\textbf{\emph{t}})p(\textbf{\emph{t}})$, we get:
\begin{small}
    \begin{equation}
    \begin{aligned}
        &D_{kl}(p(\textbf{\emph{x}})p(\textbf{\emph{t}}) , q_{\eta}(\textbf{\emph{x}} |\textbf{\emph{t}}) p(\textbf{\emph{t}}))\\
        =&\int_{\mathcal{X}}\int_{\mathcal{T}}p(\textbf{\emph{x}})p(\textbf{\emph{t}})\log\frac{p(\textbf{\emph{x}})p(\textbf{\emph{t}})}{\frac{1}{Z}p(\textbf{\emph{x}})\frac{e( t;\textbf{\emph{x}}_{t}) \phi ( z;\textbf{\emph{x}}_{z})}{e_{\eta }( t;\textbf{\emph{x}}_{t}) \phi _{\eta }( z;\textbf{\emph{x}}_{z})}p(\textbf{\emph{t}})} d\textbf{\emph{x}}d\textbf{\emph{t}}\\
        =&\int_{\mathcal{X}}\int_{\mathcal{T}}p(\textbf{\emph{x}})p(\textbf{\emph{t}})\left[ \log Z+\log\frac{e_{\eta }( t;\textbf{\emph{x}}_{t})}{e( t;\textbf{\emph{x}}_{t})}+\log\frac{\phi _{\eta } ( z;\textbf{\emph{x}}_{z})}{\phi( z;\textbf{\emph{x}}_{z})} \right]d\textbf{\emph{x}}d\textbf{\emph{t}}  \nonumber \label{equ:kl}
    \end{aligned}
\end{equation}
\end{small}
where $Z$ is the normalization factor, and according to Assumption \ref{asmp:odds ratio}, we can simplify $Z$ as follows: 
\begin{small}
    \begin{equation}
    \begin{aligned}
        Z=\int _{\textbf{\emph{x}}}p(\textbf{\emph{x}})\frac{e_{\eta }( t;\textbf{\emph{x}}_{t})\phi_{\eta } ( z;\textbf{\emph{x}}_{z})}{e( t;\textbf{\emph{x}}_{t}) \phi( z;\textbf{\emph{x}}_{z})}d\textbf{\emph{x}}\leq \int _{\textbf{\emph{x}}}p(\textbf{\emph{x}})\Gamma_t\Gamma_zd\textbf{\emph{x}}\leq \Gamma_t\Gamma_z. \label{equ:z} \nonumber
    \end{aligned}
\end{equation}
\end{small}
    
Therefore, according to Assumption \ref{asmp:odds ratio} again, we can get \\ $D_{kl}(p(\textbf{\emph{x}})p(\textbf{\emph{t}}) , q_{\eta}(\textbf{\emph{x}} |\textbf{\emph{t}}) p(\textbf{\emph{t}}))\leq 2(\log\Gamma_t+\log\Gamma_z)$. 
Then by the standard change of variables, we get:
\begin{small}
    \begin{equation}
    \begin{aligned}
        D_{kl}(p_\Phi(r)p(\textbf{\emph{t}}) , q_{_\Phi,\eta}(r |\textbf{\emph{t}}) p(\textbf{\emph{t}}))\leq 2(\log\Gamma_t+\log\Gamma_z) .\label{equ:kl bound} \nonumber
    \end{aligned}
\end{equation}
\end{small}

Next, we bound TVD between $p_\Phi(r)p(\textbf{\emph{t}})$ and $q_{_\Phi,\eta}(r |\textbf{\emph{t}}) p(\textbf{\emph{t}})$ as:
\begin{small}
    \begin{equation}
    \begin{aligned}
        \delta(p_\Phi(r)p(\textbf{\emph{t}}) , q_{_\Phi,\eta}(r |\textbf{\emph{t}}) p(\textbf{\emph{t}}))
        &\leq \sqrt{\frac{1}{2}D_{kl}(p_\Phi(r)p(\textbf{\emph{t}}) , q_{_\Phi,\eta}(r |\textbf{\emph{t}}) p(\textbf{\emph{t}}))}\\
        &\leq \sqrt{\log\Gamma_t+\log\Gamma_z},  \nonumber
    \end{aligned}
\end{equation}
\end{small}
where the first inequality follows from Pinsker's inequality. Finally, we can bound the Wasserstein distance using TVD :
\begin{small}
    \begin{equation}
    \begin{aligned}
        \mathcal{W}(p_\Phi(r)p(\textbf{\emph{t}}) || q_{_\Phi,\eta}(r |\textbf{\emph{t}}) p(\textbf{\emph{t}}))
        &\leq diam(\mathcal{R})\delta(p_\Phi(r)p(\textbf{\emph{t}}) || g_{_\Phi,\eta}(r |\textbf{\emph{t}}) p(\textbf{\emph{t}}))\\
        &\leq diam(\mathcal{R})\sqrt{\log\Gamma_t+\log\Gamma_z}, \nonumber
    \end{aligned}
\end{equation}
\end{small}
where $diam(\mathcal{R})=\sup_{r,r^{\prime}\in \mathcal{R}}\|r-r^{\prime}\|_2$, and the first inequality follows from Theorem 4 of \cite{gibbs2002choosing}. 
\end{proof}

\section{Proof of Theorem \ref{theo:tighter bound}}
\begin{proof}
Following \cite{johansson2018learning}, the results follows immediately from Lemma \ref{lem: cf bound} and the definition of IPM:
\begin{small}
    \begin{equation}
    \begin{aligned}
    \epsilon _{CF} -\epsilon _{F}^\eta &\leq B_{\Phi}\cdot IPM(p( \textbf{\emph{x}}) p(\textbf{\emph{t}}),q_\eta(\textbf{\emph{x}}|\textbf{\emph{t}})p(\textbf{\emph{t}}))\\
    &= B_{\Phi}\cdot IPM(p( \textbf{\emph{x}}) p(\textbf{\emph{t}}),w_\eta(\textbf{\emph{t}}, \textbf{\emph{x}})p(\textbf{\emph{x}}|\textbf{\emph{t}})p(\textbf{\emph{t}})).\nonumber
    \end{aligned}
\end{equation}
\end{small}

For perfect balancing weights, such as IPW, we have balancing property, which means that balancing weights can balance the reweighted distributions between different groups \cite{li2018balancing}, i.e., $p( \textbf{\emph{x}})=w(\textbf{\emph{t}}, \textbf{\emph{x}})p(\textbf{\emph{x}}|\textbf{\emph{t}})$, thus we have $IPM(p( \textbf{\emph{x}}) p(\textbf{\emph{t}}),w(\textbf{\emph{t}}, \textbf{\emph{x}})p(\textbf{\emph{x}}|\textbf{\emph{t}})p(\textbf{\emph{t}}))=0$ and the LHS is tight. 
Then by the standard change of variables, we can get Theorem \ref{theo:tighter bound}.

\end{proof}

\bibliographystyle{ACM-Reference-Format}
\bibliography{interference}

\appendix

\setcounter{proposition}{0}
\setcounter{lemma}{0}
\setcounter{theorem}{0}
\setcounter{corollary}{0}
\setcounter{equation}{0}
\numberwithin{equation}{section}

\setcounter{figure}{0}

\setcounter{section}{0}

\end{document}